\documentclass[journal]{IEEEtran}
\usepackage{cite}
\usepackage{amsmath,amssymb,amsfonts}
\usepackage{algorithmic}
\usepackage{graphicx}
\usepackage{algorithm,algorithmic}
\usepackage{multirow}
\usepackage{multicol}

\usepackage{paralist}
\usepackage{graphicx}
\usepackage{textcomp}
\usepackage{caption}

\usepackage{bm,amsmath}
\usepackage{xspace}
\usepackage{graphicx}
\usepackage{color}

\usepackage{colortbl}
\definecolor{light_gray}{RGB}{170,170,170}

\usepackage[utf8]{inputenc} 
\usepackage[T1]{fontenc}    
\usepackage{url}            
\usepackage{booktabs}       
\usepackage{amsfonts}       
\usepackage{nicefrac}       
\usepackage{microtype}      
\usepackage{longtable,lscape}
\usepackage{array,multirow}
\usepackage{enumitem}

\usepackage{wrapfig}

\newcommand{\hide}[1]{}

\begin{document}
\title{Multi-lingual Multi-institutional Electronic Health Record based Predictive Model}
\author{
Kyunghoon Hur,
Heeyoung Kwak,
Jinsu Jang,
Nakhwan Kim,
Edward Choi
\thanks{This work has been submitted to the IEEE for possible publication. Copyright may be transferred without notice, after which this version may no longer be accessible. Paper submitted for review on Dec 13, 2026.}
\thanks{Kyunghoon Hur,
 Edward Choi are with Kim Jaechul Graduate School of AI, Korea Advanced Institute of Science and Technology, Daejeon 34141, Republic of Korea (e-mail: pacesun@kaist.ac.kr; edwardchoi@kaist.ac.kr). 
}
\thanks{Heeyoung Kwak is with NAVER Digital Healthcare LAB, Republic of Korea (e-mail: heeyoung.kwak@navercorp.com).}
\thanks{Jinsu Jang, is with Department of Health and Medical Information, Ansan University, Ansan-si, Gyeonggi-do 15328, Republic of Korea (e-mail: runmc3812@ansan.ac.kr).}
\thanks{Nackhwan Kim, is with Institute of Human Behavior and Genetics, Korea University College of Medicine, 02841, Republic of Korea (e-mail: nackhwan@gmail.com). }
}
\maketitle

\begin{abstract}
Large-scale EHR prediction across institutions is hindered by substantial heterogeneity in schemas and code systems.
Although Common Data Models (CDMs) can standardize records for multi-institutional learning, the manual harmonization and vocabulary mapping are costly and difficult to scale.
Text-based harmonization provides an alternative by converting raw EHR into a unified textual form, enabling pooled learning without explicit standardization.
However, applying this paradigm to multi-national datasets introduces an additional layer of heterogeneity, which is "language" that must be addressed for truly scalable EHRs learning.

In this work, we investigate multilingual multi-institutional learning for EHR prediction, aiming to enable pooled training across multinational ICU datasets without manual standardization.
We compare two practical strategies for handling language barriers: (i) directly modeling multilingual records with multilingual encoders, and (ii) translating non-English records into English via LLM-based word-level translation. 
Across seven public ICU datasets, ten clinical tasks with multiple prediction windows, translation-based lingual alignment yields more reliable cross-dataset performance than multilingual encoders.
The multi-institutional learning model consistently outperforms strong baselines that require manual feature selection and harmonization, and also surpasses single-dataset training.

We further demonstrate that text-based framework with lingual alignment effectively performs transfer learning via few-shot fine-tuning, with additional gains.
To our knowledge, this is the first study to aggregate multilingual multinational ICU EHR datasets into one predictive model, providing a scalable path toward language-agnostic clinical prediction and future global multi-institutional EHR research. \footnote{Our code implementation is available on Github. \url{https://github.com/hoon9405/Multi-lingual-EHR-prediction}}
\end{abstract}

\begin{IEEEkeywords}
Electronic health records, natural language process, multi-institutional learning, heterogeneity, predictive model, multi-lingual, multi-task learning. 
\end{IEEEkeywords}

\begin{figure}[ht] 
    \includegraphics[width=.50\textwidth]{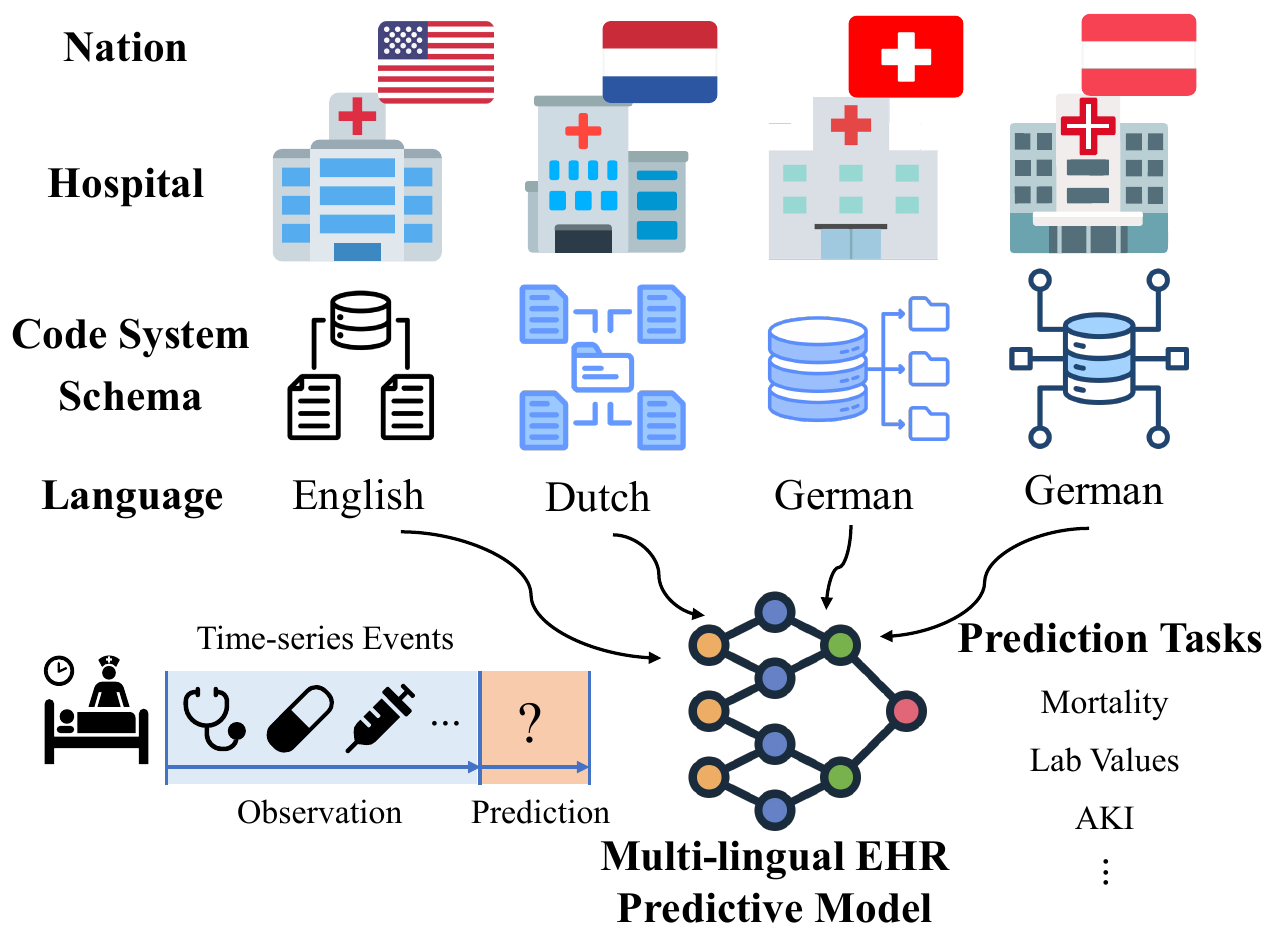}
    \caption{
    Overview of the multilingual multi-institutional EHR predictive framework.
    EHR time-series from multiple hospitals and nations—recorded under heterogeneous code systems, schemas, and languages (e.g., English, Dutch, German)—are harmonized into a unified textual/event representation and jointly used to train a single foundational predictive model.
    The resulting model is evaluated across diverse ICU prediction tasks, demonstrating scalable pooled learning without manual standardization.
    }
    \label{fig:fig1}
    \vskip -4pt
\end{figure}

\section{Introduction}
\label{sec:introduction}
\IEEEPARstart{E}{lectronic} Health Record (EHR) plays a central role in modern clinical research, providing structured and unstructured information such as diagnoses, medications, laboratory results, and procedures.
Such comprehensive clinical data enable the development of artificial intelligence (AI) models capable of predicting high-risk conditions, including cardiovascular events, sepsis, and acute kidney injury (AKI), before they manifest clinically~\cite{choi2016doctor,awad2017early,thiel2010early,shameer2017predictive}.
These predictive capabilities not only improve patient safety through timely intervention, but also help reduce healthcare costs and optimize the use of scarce medical resources~\cite{rajkomar2018scalable,meystre2017secondaryuse}.

Despite these advances, existing EHR-based predictive model faces several challenges.
Many previous studies have focused narrowly on specific subsets of patients (e.g., undergoing specialized procedures), which are from a single institution and limit the applicability of these models to the broader inpatient population with various clinical needs~\cite{hyland2020circfailure,thoral2021amsterdamumcdb_ml}.
As deep learning continues to evolve, leveraging multi-institutional datasets has become essential to improve model robustness, generalizability, and external validity~\cite{sauer2022publicicu,rajkomar2018scalable}.
However, multi-institutional research remains difficult due to substantial \textit{heterogeneity} across EHR systems, including differences in data schemas, coding practices, and documentation styles~\cite{meystre2017secondaryuse,hripcsak2015ohdsi}.
Harmonizing features across institutions requires extensive and often manual preprocessing, imposing significant time and resource burdens.

To address this heterogeneity, many collaborative initiatives have adopted CDMs, such as the OMOP, which provide a unified schema for cross-institutional analysis~\cite{hripcsak2015ohdsi,park2019ohdsi}.
While CDMs offer a structured foundation for multi-site research, implementing and maintaining them requires considerable institutional investment~\cite{park2019ohdsi,belenkaya2021extending}.
OMOP transformation typically involves clinicians, data managers, and database engineers, and the process may introduce mapping errors or information loss~\cite{belenkaya2021extending}.
These barriers pose substantial challenges, particularly for smaller institutions with limited resources, and ultimately constrain the scalability of predictive modeling.

Recently, text-based approaches have emerged as a promising alternative for processing EHR data without relying on such rigid standardization frameworks~\cite{hur2024genhpf, hur2022descemb, singhal2023llmclinic}.
By directly interpreting raw or minimally processed clinical text, text-based approaches can reduce the need for handcrafted feature engineering and accommodate diverse EHR structures~\cite{yang2022gatortron}.
This flexibility suggests that text-based approach may offer a scalable solution for cross-institutional predictive modeling.
However, existing text-based approaches have primarily been validated on English-language datasets such as MIMIC-III~\cite{johnson2016mimic3} and eICU~\cite{pollard2018eicu}.
In reality, global healthcare systems operate in numerous languages, and the generalizability of text-based frameworks to non-English EHR environments remains largely unexplored.
Overcoming this linguistic dependency is essential for developing medical AI systems that are broadly deployable across countries and healthcare settings.

In this study, we aim to address this gap by empirically examining how an existing text-based EHR predictive framework performs when extended to multilingual, multi-institutional EHR datasets.
Using publicly available datasets, including Dutch and German~\cite{thoral2021amsterdamumcdb,faltys2021hirid}, we adapt text-based EHR predictive framework~\cite{hur2024genhpf, kim2024remed} to a setting where heterogeneous EHR tables are converted into a unified textual representation without any harmonization process.
Figure~\ref{fig:fig1} provides an overview of this setup.
Within this setup, we systematically explore two practical strategies for handling multilingual EHR data.
The first approach leverages multilingual models capable of jointly encoding records across different languages.
The second approach takes advantage of the fact that many non-English clinical datasets contain a substantial portion of English terminology; thus, we unify multilingual inputs by applying LLM-based word-level translation into English and subsequently train a single model using the text-based framework.

Through empirical evaluation, we find that the latter approach—LLM-assisted translation followed by text-based model consistently outperforms all other baseline models for EHR prediction tasks when multi-institutional learning cases.
Our experiments span seven heterogeneous ICU cohorts and consider mult-institutional learning case and multilingual configurations.

The main contributions of our work can be summarized as follows:
\begin{enumerate}
    \item \textbf{First multilingual, multi-institutional prediction with publicly available EHRs.}  
    Using seven publicly available EHR datasets from different countries and languages, we train text-based EHR models and evaluate performance separately on each site across ten clinical prediction tasks and multiple prediction windows.
    To our knowledge, this is among the first studies to perform pooled multilingual, multi-institutional prediction across ICU cohorts drawn from different national health systems using a single text-based modeling framework.

    \item \textbf{Cross-language few-shot transfer.}  
    We investigate cross-language transfer by adapting models trained on one or more source languages to a target dataset with limited labeled data, including settings where the target is translated into a shared English space.
    These experiments show that text-based models on linguistically aligned EHR can achieve strong performance in a few-shot scenarios, providing a practical way to extend models to EHR data with different languages.

    \item \textbf{Comparison of multilingual modeling settings and manually aligned feature baselines.}  
    We compare several strategies for multilingual, multi-institutional learning with EHR data, including feature-based baselines constructed with domain-driven variable selection for each dataset, multilingual encoders trained directly on multilingual text, and translation-based text models that map heterogeneous languages into a unified English representation.
    Applying text-based model with lingually aligned EHR trained on pooled multi-lingual datasets consistently outperforms both single-dataset trained models and the feature aligned baselines.
\end{enumerate}

\renewcommand{\arraystretch}{1.3}
\begin{table*}[ht]
    \centering
    \begin{tabular}{cccccccc}
    \toprule
    Datasets & MIMIC-IV & eICU & NWICU & EHRShot & UMCdb & HiRID & SICdb \\
    \midrule
    Nation   & USA & USA & USA & USA & Netherlands & Switzerland & Austria \\
    Language & English & English & English & English & Dutch & German & German \\
    \midrule
    No. of Observations           & 65511 & 98904 & 42137 & 6739  & 23182 & 33905 & 27461 \\
    No. of ICU stays              & 65511 & 98904 & 42137 & 6739  & 23182 & 33905 & 27461 \\
    Mean No. of events per sample & 88.7  & 48.5  & 74.8  & 247.6 & 118.9 & 149.3 & 111.4 \\
    \midrule
    No. of Unique code                     & 9565 & 6302 & 8173 & 11942 & 9041 & 10087 & 8526 \\
    No. of Unique subwords text            & 3512 & 2678 & 2976 & 4037  & 3189 & 3348  & 3097 \\
    Mean No. features per event            & 10.0 & 6.7  & 8.4  & 9.1   & 9.6  & 11.2  & 8.9 \\
    Mean length of subwords text per event & 62.2 & 51.0 & 54.7 & 60.8  & 57.5 & 59.6  & 56.9 \\
    \bottomrule
    \end{tabular}
    \caption{\label{tab:tabstat} Statistics of datasets}
    \vskip -10pt
\end{table*}

\section{Related Work} \label{sec:related work}

\noindent{\textbf{Multi-institutional EHR learning}}
Multi-institutional EHR learning has often relied on standardization such as OMOP~\cite{hripcsak2015ohdsi}, which requires each site to manually convert its EHR into a shared schema.
Although OMOP enables cross-site studies, the overhead of Extract, Transform, Load (ETL) and vocabulary mapping limits scalability.
To ease this burden, MEDS (Medical Event Data Standard)~\cite{arnrich2024meds} defines an event-centric minimal format and accompanying tools so heterogeneous public datasets can be processed through a unified pipeline.
Yet Another ICU Benchmark(YAIB)~\cite{vandeWater2023yaib} builds on this idea by standardizing common variables across ICU datasets by benchmarking pooled and transfer learning.
Still, both assume alignment to shared formats/features and do not address pooled learning over multilingual EHR without explicit standardization.

An alternative privacy-preserving approach is to share models rather than raw data.
Guo et al.~\cite{guo2024ehrfm} pretrained a structured EHR foundation model at one institution and adapted it to external sites through continued pretraining or fine-tuning.
Although effective for transfer, this differs from our setting that jointly trains on multiple institutions and languages from the outset.

\noindent{\textbf{Resolving EHR heterogeneity with text-based approach}} 
A major obstacle to multi-institutional learning is heterogeneity in codes and schemas.
Graph-based alignment methods aim to reduce this gap: MIKGI~\cite{zhou2022mikgi} models site-specific codes as incomplete multi-view knowledge graphs and learns shared embeddings by combining code descriptions with co-occurrence statistics, enabling cross-site code alignment, but still remains code-level.

Text-based harmonization offers another route.
DescEmb~\cite{hur2022descemb} embeds natural-language code descriptions into a shared semantic space, reducing manual code mapping but still requiring site-specific feature preprocessing.
GenHPF~\cite{hur2024genhpf} goes further by converting heterogeneous tables into hierarchical text sequences, resulting in a schema-agnostic framework for pooled/transfer learning; however, it has been validated mainly in English-only EHR settings.

\begin{figure}[ht] 
    \centering
    \includegraphics[width=1\linewidth]{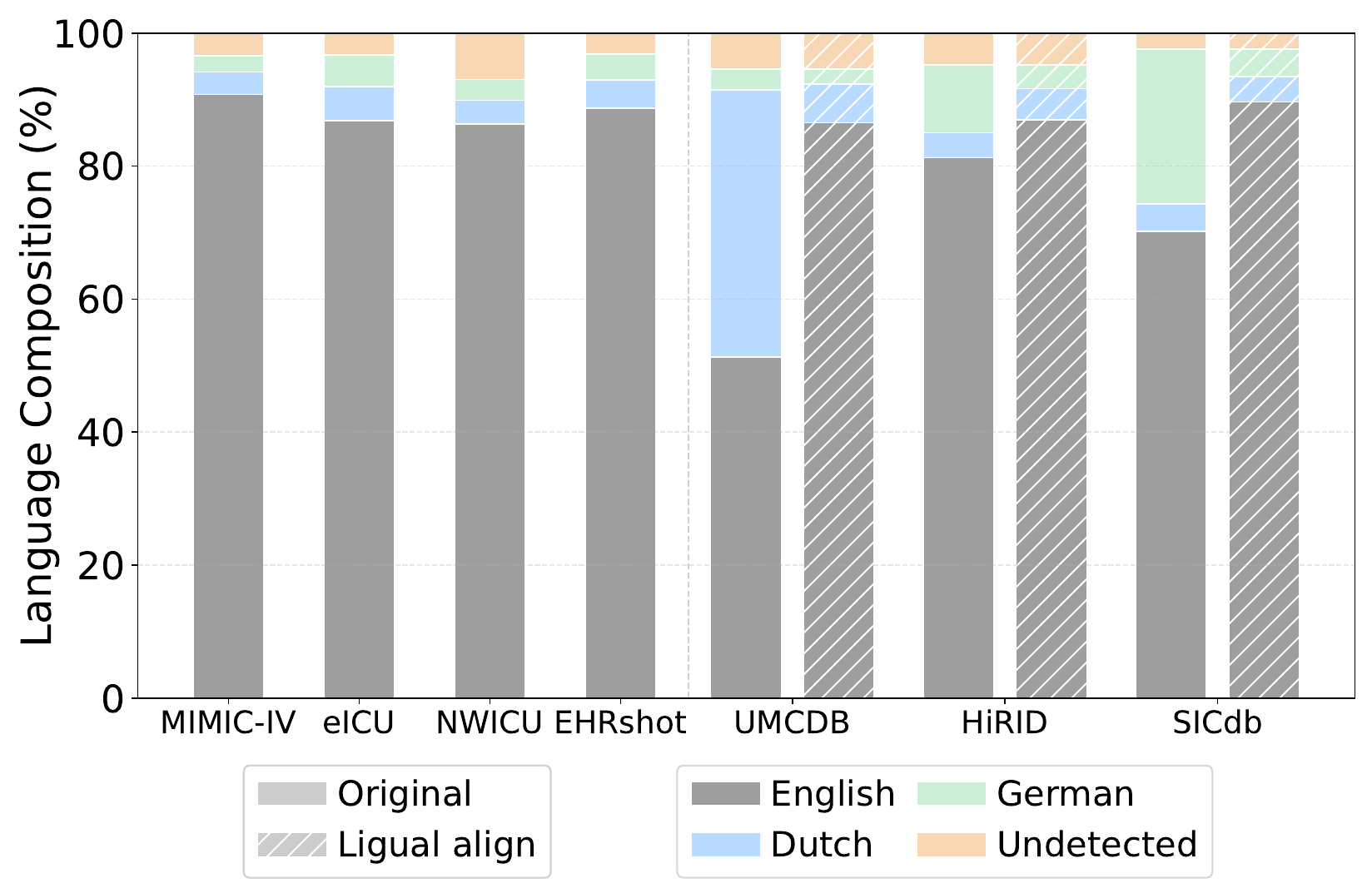}
    \caption{Language composition of clinical text across seven ICU EHR datasets. Each bar shows the percentage of tokens identified as English (en), Dutch (nl), German (de), or undetected by our language identification pipeline. “Undetected” denotes tokens that are not present in standard word lexicons and are not confidently assigned to any language by the identifier—typically domain-specific abbreviations or proper nouns (e.g., “PO” for oral administration). 
    }
    \vskip -4pt
    \label{fig:fig3}
\end{figure}

\noindent{\textbf{Clinical multi-lingual encoders and translation}}
BERT~\cite{devlin2019bert} provides a general-purpose English encoder for clinical text, and domain-adapted variants such as BioBERT~\cite{lee2020biobert} and ClinicalBERT~\cite{alsentzer2019clinicalbert} improve biomedical and clinical understanding by continuing pretraining on biomedical literature and clinical notes.

Because these encoders are English-specific, multilingual pretrained language models (PLMs) such as mBERT~\cite{pires2019mbert} and XLM-Roberta\cite{conneau2020xlmr} are commonly used for cross-lingual transfer.
More specialized biomedical/clinical multilingual encoders, including KBioXLM~\cite{geng2023kbioxlm} (English–Chinese) and CLIN-X~\cite{lange2022clinx} (English–Spanish), further adapt these models to biomedical or clinical corpora, but evaluations have focused on NLP tasks rather than multilingual ICU prediction.

Another practical strategy is translation-based unification: non-English records are translated into English with LLMs and then modeled with English clinical encoders.
We adopt this approach and compare it directly against multilingual encoders for EHR prediction.

\begin{figure*}[ht]
    \includegraphics[width=\textwidth]{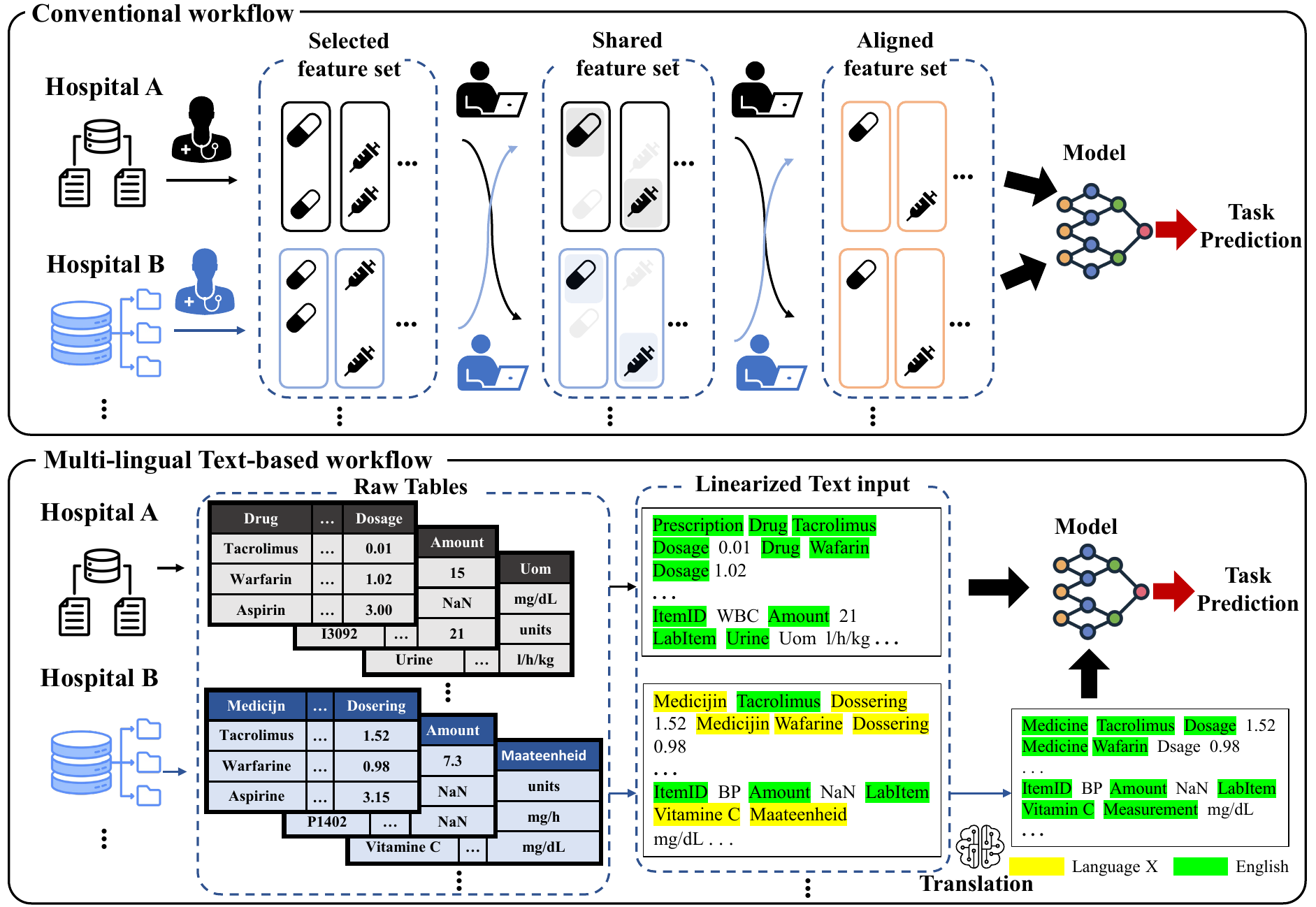}
    \caption{
    Comparison of conventional multi-institutional learning and our multi-lingual text-based workflow.
    The conventional pipeline (top) requires each hospital to manually select, share, and align a common feature set across heterogeneous EHR schemas and code systems before model training.
    In contrast, the proposed workflow (bottom) linearizes raw EHR tables from different institutions and languages into a unified textual representation, applies LLM-based translation to standardize non-English content into English, and trains a single pooled model directly on the harmonized text for downstream task prediction with minimal preprocessing.}
    \label{fig:fig2}
    \vskip -4pt
\end{figure*} 

\section{Methodology} \label{sec:math}
\subsection{Datasets}

We use seven publicly available EHR datasets that span multiple institutions, countries, and languages:
\begin{itemize}
    \item \textbf{MIMIC-IV}~\cite{johnson2021mimic}: a large single-center ICU database from Beth Israel Deaconess Medical Center (US), containing de-identified demographics, diagnoses, medications, labs, and bedside measurements.
    \item \textbf{eICU-CRD}~\cite{pollard2018eicu}: a multi-center ICU dataset from tele-ICU programs in the US, covering various hospital sites with high-granularity clinical events.
    \item \textbf{NWICU}~\cite{nwicu2024}: a contemporary ICU cohort from Northwestern Memorial HealthCare (US), enriched with recent clinical practice patterns and COVID-era care.
    \item \textbf{EHRSHOT}~\cite{ehrshot2023}: a longitudinal structured EHR benchmark from Stanford Medicine (US), not restricted to ICU stays, included to broaden institutional diversity.
    \item \textbf{AmsterdamUMCdb (UMCdb)}~\cite{thoral2021amsterdamumcdb}: an ICU dataset from the Netherlands, representing European clinical documentation and workflows.
    \item \textbf{HiRID}~\cite{yeche2021hirid}: a high-resolution ICU dataset from Bern University Hospital (Switzerland), featuring dense vital-sign and lab trajectories.
    \item \textbf{SICdb}~\cite{rodemund2023sicdb}: an ICU dataset from University Hospital Salzburg (Austria), providing additional non-English, multi-national ICU records.
\end{itemize}

Across all datasets, we extract all time-stamped medical events such as laboratory tests, medications, procedures, vital signs, clinical observations, etc, which serve as model inputs.
We construct adult cohorts by selecting patients aged $\ge 18$ yearswith an ICU length of stay of at least 24 hours, consistent with prior EHR benchmarks~\cite{rajkomar2018scalable, vandeWater2023yaib}. 
Each dataset is split into training, validation, and test sets at the patient level in an 8:1:1 ratio to ensure unbiased evaluation. Figure~\ref{fig:fig3} summarizes the language distribution in each dataset with original dataset and dataset with translation to English.
Table~\ref{tab:tabstat} summarizes the general statistics of the seven datasets.

We follow the minimal schema-agnostic preprocessing pipeline introduced in GenHPF and ReMed~\cite{hur2024genhpf, kim2024remed} for all datasets.
In brief, preprocessing consists of three steps: (i) removing non-informative identifier-like fields (e.g., columns containing only integers), (ii) representing numeric values using digit-place embedding, and (iii) converting remaining structured events into hierarchical textual sequences and tokenizing them with a various version of tokenizer.

\subsection{EHR predictive model}
For each institution $s \in \mathcal{S}$, a patient $P$ is associated with an EHR record
$\mathbf{R}^{(s)}_P = [\mathcal{M}_1,\ldots,\mathcal{M}_N]$,
consisting of time-ordered medical events observed within a predefined input window.
Each event $\mathcal{M}_i$ comprises a set of name-value feature pairs
$\{(n_i^{k}, v_i^{k})\}$, an event type $e_i$ (e.g., lab, medication), and an associated timestamp $T_i$.

Given a clinical prediction setting, the model uses events in the input window to predict
an outcome $y$ within a predefined prediction horizon.
The objective is to learn a predictor that maps $\mathbf{R}_P$ to $\hat{y}$.
For simplicity, we omit the site index when the context is clear and write $\mathbf{R}_P$ instead of $\mathbf{R}^{(s)}_P$.

\subsection{Text-based embedding for multi-institutional learning}
Given the EHR representation described above, the remaining challenge for multi-institutional learning is to construct event representations that can be shared across institutions with heterogeneous schemas and local code systems.
Rather than relying on site-specific code vocabularies or manual alignment,
we adopt a text-based embedding approach that represents each medical event directly as text (Figure~\ref{fig:fig2}) following GenHPF~\cite{hur2024genhpf}.

For an event $\mathcal{M}_i$, we linearize its content into a textual sequence
\begin{equation}
u_i = \mathrm{Lin}\!\big(e_i, \{(n_i^{k}, v_i^{k})\}\big),
\label{eq:linearize}
\end{equation}
where $\mathrm{Lin}(\cdot)$ deterministically serializes the event type together with all associated feature name-value pairs.
This operation preserves the full event information while producing a unified text representation that is independent of local schemas or coding conventions.

A shared tokenizer $S$ and text encoder $f$ are applied to obtain event embeddings, which are then
aggregated into a patient representation for prediction:
\begin{align}
\mathbf{m}_i &= f\!\left(S(u_i)\right), \label{eq:text_event_embed}\\
\mathbf{h}_P &= g\!\left([\mathbf{m}_1,\ldots,\mathbf{m}_N]\right), \label{eq:patient_embed}\\
\hat{y}      &= h\!\left(\mathbf{h}_P\right), \label{eq:prediction_head}
\end{align}
where $g$ denotes an event-sequence encoder and $h$ a prediction head, both trained end-to-end with $f$.

Here, $\mathrm{Lin}(\cdot)$ converts each structured event into a text sequence $u_i$,
$S(\cdot)$ is a shared subword tokenizer, and $f(\cdot)$ is a text encoder that maps the tokenized event into an event embedding $\mathbf{m}_i$.
The sequence encoder $g(\cdot)$ aggregates the event embeddings into a patient representation $\mathbf{h}_P$,
and the prediction head $h(\cdot)$ produces the final output $\hat{y}$ (e.g., classification probabilities or a real-valued target).
All modules are trained end-to-end.

\begin{table*}[ht]
\centering
\renewcommand{\arraystretch}{1.5}
{\setlength{\tabcolsep}{4.5pt}
\begin{tabular}{cc c c c c c c c c}
\toprule
\multirow{2}{*}{Dataset} & \multirow{2}{*}{Train mode}
& \multicolumn{1}{c}{YAIB*}
& \multicolumn{1}{c}{Rajikomar*}
& \multicolumn{3}{c}{GenHPF}
& \multicolumn{3}{c}{ReMed} \\
\cmidrule(lr){3-3}\cmidrule(lr){4-4}\cmidrule(lr){5-7}\cmidrule(lr){8-10}
& &
\multicolumn{1}{c}{Code-base} &
\multicolumn{1}{c}{Code-base} &
\multicolumn{1}{c}{EnTok} &
\multicolumn{1}{c}{MLTok} &
\multicolumn{1}{c}{LLM Align} &
\multicolumn{1}{c}{EnTok} &
\multicolumn{1}{c}{MLTok} &
\multicolumn{1}{c}{LLM Align} \\
\midrule

\multirow{2}{*}{MIMIC-IV}
& Single & \textbf{0.782} & 0.775 & 0.767 & 0.767 & 0.767 & 0.777 & 0.774 & 0.777 \\
& Multi  & 0.770 & 0.768 & 0.771 & 0.770 & 0.779 & 0.780 & 0.780 & \textbf{0.783} \\
\cmidrule(lr){1-10}

\multirow{2}{*}{eICU}
& Single & \textbf{0.774} & 0.760 & 0.754 & 0.754 & 0.754 & 0.771 & 0.767 & 0.771 \\
& Multi  & 0.757 & 0.753 & 0.761 & 0.761 & 0.775 & 0.772 & 0.773 & \textbf{0.780} \\
\cmidrule(lr){1-10}

\multirow{2}{*}{EHRshot}
& Single & \textbf{0.814} & 0.805 & 0.802 & 0.798 & 0.802 & 0.812 & 0.808 & 0.812 \\
& Multi  & 0.802 & 0.795 & 0.812 & 0.807 & \textbf{0.820} & 0.815 & 0.817 & 0.817 \\
\cmidrule(lr){1-10}

\multirow{2}{*}{NWICU}
& Single & \textbf{0.757} & 0.744 & 0.744 & 0.740 & 0.744 & 0.754 & 0.752 & 0.754 \\
& Multi  & 0.749 & 0.743 & 0.752 & 0.747 & 0.760 & 0.766 & 0.761 & \textbf{0.770} \\
\cmidrule(lr){1-10}

\multirow{2}{*}{SICdb}
& Single & 0.756 & 0.747 & 0.745 & 0.741 & 0.747 & \textbf{0.757} & 0.753 & 0.756 \\
& Multi  & 0.746 & 0.737 & 0.748 & 0.748 & 0.761 & 0.763 & 0.758 & \textbf{0.765} \\
\cmidrule(lr){1-10}

\multirow{2}{*}{UMCdb}
& Single & \textbf{0.761} & 0.742 & 0.744 & 0.738 & 0.745 & 0.755 & 0.751 & 0.757 \\
& Multi  & 0.752 & 0.736 & 0.748 & 0.746 & 0.759 & 0.757 & 0.756 & \textbf{0.762} \\
\cmidrule(lr){1-10}

\multirow{2}{*}{HiRiD}
& Single & \textbf{0.813} & 0.803 & 0.806 & 0.800 & 0.808 & 0.807 & 0.806 & 0.805 \\
& Multi  & 0.808 & 0.795 & 0.811 & 0.807 & 0.813 & 0.811 & 0.810 & \textbf{0.813} \\

\hline\hline
\multirow{2}{*}{Avg}
& Single & \textbf{0.780} & 0.768 & 0.766 & 0.763 & 0.767 & 0.776 & 0.773 & 0.776 \\
& Multi  & 0.774 & 0.765 & 0.777 & 0.774 & 0.785 & 0.784 & 0.783 & \textbf{0.788} \\
\bottomrule
\end{tabular}
}
\caption{Multi-institutional learning results on seven ICU datasets. For each dataset, AUROC values are averaged over all prediction tasks and five random seeds under two training regimes: \emph{Single} (trained on one dataset) and \emph{Multi} (one model jointly trained on the union of all datasets and evaluated per site). YAIB* and Rajkomar* are code-based baselines, while GenHPF and ReMed are schema-agnostic text-based frameworks evaluated with different tokenization/alignment strategies (EnTok: English-only tokenizer; MLTok: multilingual tokenizer; LLM Align: aligned to EnTok). Bold values indicate the best performance within each dataset and training regime; \emph{Avg} reports the mean across datasets.}
\label{tab:multi}
\end{table*}

\subsection{Multi-lingual multi-institutional prediction model}\label{subsec:multilingual_model}
Building on text-based event representation in Eq.~\ref{eq:linearize}--\ref{eq:text_event_embed},
we pool heterogeneous EHR datasets across institutions and languages to train a single shared model.
Let $\ell \in \mathcal{L}$ denote the language of an EHR dataset, and let $\mathcal{D}$ denote the pooled
collection of patient records across all datasets.
Our goal is to learn a shared parameter set $\theta$ that produces transferable patient representations
across institutions, without requiring language- or dataset-specific model components.

\noindent\textbf{Two strategies for multilingual modeling.}
We consider two ways to obtain language-compatible event embeddings.

\emph{(i) Multilingual-encoder pooled modeling.}
Each linearized event string $u_i$ is tokenized and embedded using a multilingual tokenizer
$S_{\mathrm{multi}}$ and encoder $f_{\mathrm{multi}}$:
\begin{equation}
\mathbf{m}_i^{\mathrm{multi}} = f_{\mathrm{multi}}\!\left(S_{\mathrm{multi}}(u_i)\right).
\label{eq:multi_encoder}
\end{equation}
This yields a shared embedding space by design, enabling pooled training directly on multilingual text.

\emph{(ii) Translation-based unification.}
Alternatively, we map each event string into a common language (English) using an LLM-based word-level translation operator $\mathcal{T}$:
\begin{equation}
\tilde{u}_i = \mathcal{T}(u_i), \qquad \tilde{u}_i \in \text{English}.
\label{eq:unify_translate}
\end{equation}
A shared English tokenizer $S_{\mathrm{en}}$ and encoder $f_{\mathrm{en}}$ are then applied:
\begin{equation}
\mathbf{m}_i^{\mathrm{en}} = f_{\mathrm{en}}\!\left(S_{\mathrm{en}}(\tilde{u}_i)\right).
\label{eq:en_encoder}
\end{equation}
Because all inputs are unified into the same language, this approach allows us to leverage strong English encoders while preserving the schema-agnostic text workflow.
We compare both strategies empirically, and use the translation-based strategy as the default setting in our experiments.

\subsection{Implementation details}
\subsubsection{Prediction tasks}
We evaluate our multilingual, multi-institutional framework on ICU-stay–level prediction tasks adapted from YAIB~\cite{vandeWater2023yaib}.
We include all ICU stays in a single cohort (no cohort stratification as in YAIB).
If a label is not available for a stay (e.g. missing measurements or dialysis-related filtering), we mask that task for the stay and do not propagate loss.

\begin{compactitem}
    \item \textit{Mortality} (binary): $k$-day mortality for $k \in \{1,2,3,7,14\}$. A stay is positive if death occurs within $k$ days after the observation window.
    \item \textit{Length of stay (LOS)} (binary): $k$-day LOS for $k \in \{7,14\}$. A stay is positive if the remaining ICU stay exceeds $k$ days.
    \item \textit{Laboratory values} (multi-class): future lab severity prediction for $k \in \{1,2,3\}$ days. We discretize the most recent value within each horizon into clinically defined categories. 
    Targets include creatinine, platelets, WBC, hemoglobin, bicarbonate, sodium, and urine output.
    Detailed descretization criteria is described in Appendix~\ref{apd:data}.
    \item \textit{Acute kidney injury (AKI)} (multi-class): $k$-day AKI staging for $k \in \{1,2,3\}$ days using KDIGO-like rules~\cite{kdigo2012aki} (with dialysis indicators when available), producing ordinal stages following YAIB~\cite{vandeWater2023yaib}.
\end{compactitem}

Binary tasks are evaluated with AUROC and multi-class tasks use one-vs-rest AUROC.
Using medical event information from the initial 12 hours after ICU admission, we apply a 12-hour timegap across all tasks.
Excluding any ICU stays shorter than 24 h allows for both a 12-h observation window and a 12-h gap.
Statistics for prediction labels are shown in Tables~\ref{tab:binary-class} and ~\ref{tab:multi-class}.

\subsubsection{Baselines}
To our knowledge, no prior study has evaluated multilingual, multi-institutional EHR prediction at scale under a unified experimental setting.
Therefore, we compare our approach against representative multi-institutional EHR learning frameworks that reflect different design choices on (i) how heterogeneity is handled (manual standardization vs. text-based schema-agnostic modeling) and (ii) how multilinguality is treated (native multilingual encoders vs. translation-based unification).
All baselines are trained and evaluated on the same seven datasets, tasks, and prediction windows, using identical input but model-specific embedding approaches to ensure a fair comparison.

\begin{compactitem}
\item YAIB*~\cite{vandeWater2023yaib}: A benchmark-driven multi-institutional baseline that performs pooled learning over multiple ICU datasets using a predefined set of common variables/features.
This baseline represents the standard practice of aligning datasets by selecting overlapping features and training one model across sites.
We use only dynamic feature sets pre-defiend in YAIB except static features in this study.

\item Rajkomar*~\cite{rajkomar2018scalable}: A multi-task EHR prediction framework that utilizes all available features within each dataset.
In our implementation, each structured feature is treated as a discrete code token, and continuous variables are discretized into value bins, so that are represented within a code-based vocabulary.
We modified Rajikomar to hierarchical structure like GenHPF model architecture but using code-based approach.

\item GenHPF~\cite{hur2024genhpf}: A text-based, schema-agnostic predictive framework that converts heterogeneous EHR tables into hierarchical textual sequences, enabling pooled and transfer learning without manual code or schema harmonization. 

\item ReMed~\cite{kim2024remed}: A text-based EHR prediction framework that, like GenHPF, converts heterogeneous records into textual representations but assumes an effectively infinite observation window over all past encounters.
ReMed learns to retrieve and aggregate the most relevant past codes for each prediction..

\end{compactitem}



\subsubsection{Model implementation}
All methods are trained on the same time-stamped medical events of each ICU stay.
GenHPF and ReMed are implemented following the original frameworks and architectures.  
ReMed allows its retriever to access all events prior to the prediction time, including history before ICU admission~\cite{kim2024remed}.

YAIB uses the 48 common variables defined in the original benchmark except demographic information; for datasets not included in YAIB~\cite{vandeWater2023yaib} (EHRSHOT~\cite{ehrshot2023}, SICdb~\cite{rodemund2023sicdb}, NWICU~\cite{nwicu2024}), we manually mapped features to these variables so that all seven datasets uses the same 48-dimensional time-series feature set. 
The Rajkomar~\cite{rajkomar2018scalable} baseline adopts a code-based representation structurally similar to GenHPF~\cite{hur2024genhpf} but does not tokenize text: each unique clinical code is treated as a categorical feature with its own embedding, and continuous values are discretized into categorical features.

For our multilingual experiments, we evaluate a translation-based text representation setting built on existing text-based EHR frameworks (GenHPF and ReMed).
Specifically, we use the same time-stamped medical events as GenHPF/ReMed, translate non-English descriptions into English with an LLM, serialize each event into hierarchical English text, and then tokenize/encode all datasets with a single English tokenizer and a shared Transformer backbone~\cite{hur2024genhpf,devlin2019bert}.

\subsubsection{Lingual alignment}

We align Dutch and German tokens to a shared English space in two steps: token-level language identification followed by word-level translation.

\paragraph{Token-level language identification.}
We first split all clinical text fields into whitespace-separated tokens and assign a language label to each token.
For tokens that match exactly one of three lexicons (English, Dutch, German) constructed from public word lists, we directly assign that language.
The remaining tokens are classified with \texttt{langid}~\cite{lui2012langid}, an off-the-shelf language identification tool trained on 97 languages; we keep labels only when they fall into {en, nl, de}.
Tokens that are still unresolved are passed to GPT-4 together with their local context, asking which of the three languages they most likely belong to.
If the model cannot confidently decide, we mark the token as \textit{undetected}, which typically corresponds to abbreviations or proper nouns (e.g., PO'' or SpO\textsubscript{2}'').

\paragraph{Word-level translation.}
Translation is applied only to tokens identified as Dutch or German; English and \textit{undetected} tokens are left unchanged.
We consider two alignment strategies.
The dictionary-based variant uses bilingual lexicons (English--Dutch, English--German, Dutch--German) complemented with a small medical term dictionary to replace non-English tokens with their English counterparts when available; otherwise the original token is kept.
The LLM-based variant instead queries Qwen3-Instruct-8B and Medgemma-7b-it with each non-English token and its surrounding feature text, and uses the returned English term when a clear answer is provided.
We report ablation results for these LLM choices in Table~\ref{tab:ablation}, and provide additional implementation details in the Appendix~\ref{apd:prompt}.
We apply these procedures to the non-English ICU datasets (UMCdb, SICdb, HiRID), while MIMIC-IV and eICU remain predominantly English but contain occasional Dutch or German terms due to local drug or device names.
After alignment, all text is serialized into English sentences and tokenized with a single English tokenizer for downstream modeling.

\begin{figure*}[ht] 
    \includegraphics[width=1\textwidth]{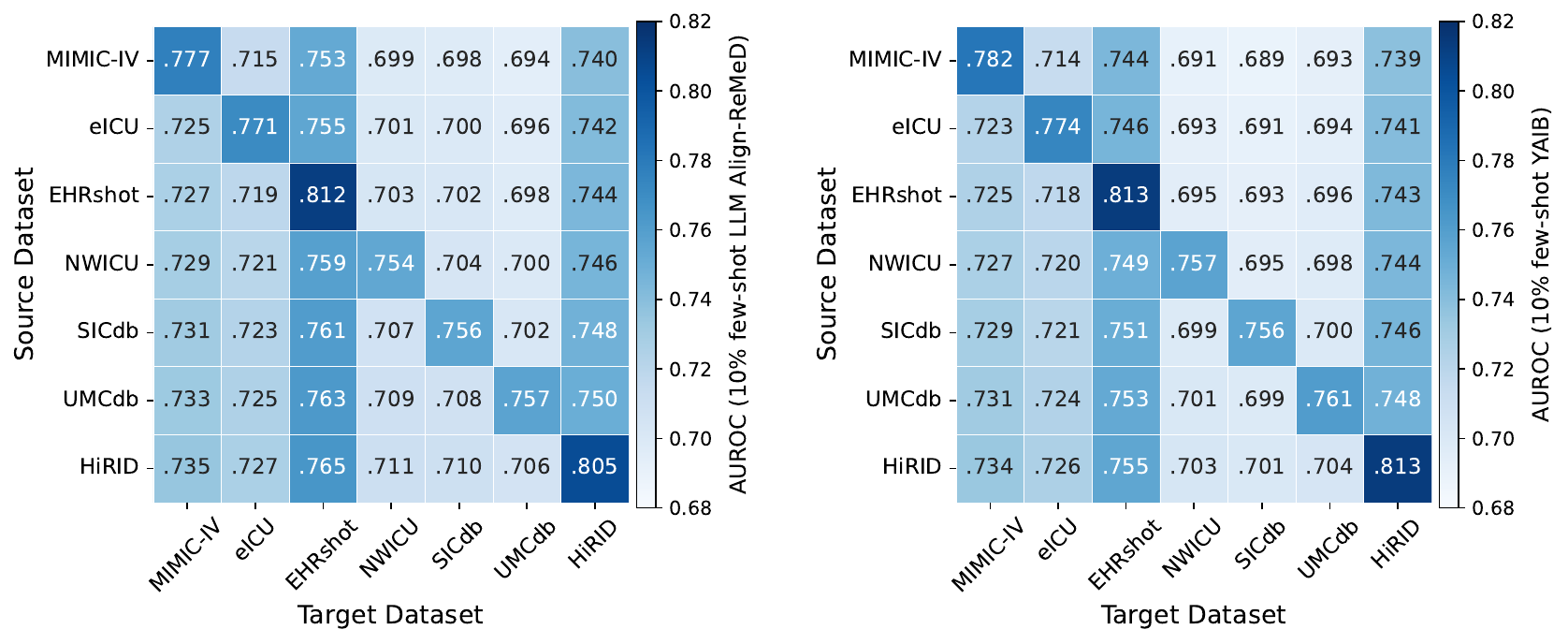}
    \caption{
    Cross-site transfer learning performance across seven datasets when fine-tuning on only 10\% of the target data for ReMED+LLM Align (left) and YAIB (right). Each heatmap shows AUROC when training on the source dataset (rows) and evaluating on the target dataset (columns); diagonal cells represents the single-dataset models (no transfer) , while off-diagonal cells correspond to transfer learning via fine-tuning on the target site. 
    }
    \label{fig:fig4}
    \vskip -4pt
\end{figure*}

\subsubsection{Training details}
All experiments are conducted using five random seeds which are used to initialize the model parameters and to split the dataset.
Their performance is evaluated based on the area under the receiver operating characteristics (AUROC) averaged over twelve tasks.
We conduct all experiments in a multi-task learning setting, as our main interest is to develop a single model that performs multiple tasks using multiple EHR datasets simultaneously.
For multi-institutional learning, we train the combined dataset and validate each individual dataset separately.
Early stopping is enforced according to the validation AUROC for each dataset, and the best model is saved per dataset. 
Subsequently, each saved model is used to test the corresponding dataset.

\section{Experiments} \label{sec:results}

We conduct experiments to evaluate multilingual, multi-institutional EHR prediction under heterogeneous schemas, coding systems, and documentation languages.
Our evaluation focuses on three aspects: (i) the effect of pooling heterogeneous datasets during training, (ii) different strategies for handling multilingual clinical text within schema-agnostic models, and (iii) the adaptability of pooled models to new institutions with limited labeled data.

\subsection{Evaluation for multi-institutional learning}

We first evaluate whether pooling heterogeneous ICU EHR datasets into a single model improves predictive performance at individual institutions, and how this effect depends on the representation and language-handling strategy.
Although pooled training is commonly assumed to improve generalization through increased data diversity, its effectiveness in EHR settings is nontrivial due to substantial heterogeneity across institutions.

We compare two code-based baselines (YAIB*, Rajkomar*), which rely on explicitly aligned feature sets, with two schema-agnostic text-based frameworks (GenHPF, ReMed), which operate directly on linearized event text.
For the text-based models, we consider three multilingual configurations: (i) an English-only tokenizer applied directly to the original text (EnTok), (ii) a multilingual tokenizer and encoder trained on mixed-language inputs (MLTok), and (iii) LLM-based word-level alignment of non-English tokens into English followed by English tokenization (LLM Align).

To isolate the effect of multi-institutional pooling, we evaluate two training regimes.
In the \textit{Single} setting, a separate model is trained and evaluated on each dataset independently.
In the \textit{Multi} setting, a single shared model is trained on the union of all seven datasets and evaluated separately on each site.
All models are trained in a multi-task setting using identical prediction tasks, input windows, and optimization procedures.
Performance is measured using AUROC, averaged across all tasks and five random seeds.

Table~\ref{tab:multi} reports per-dataset and average performance under both training regimes.
Under \textit{Single} training, YAIB achieves the strongest average performance across datasets, while GenHPF and ReMed achieve comparable results without relying on manual feature alignment.
Under \textit{Multi} training, the text-based models benefit more consistently from pooled learning than the code-based baselines.
Among all configurations, LLM-based alignment combined with an English tokenizer yields the highest average AUROC, indicating that reducing linguistic variability facilitates more effective cross-institutional sharing.

\subsection{Transfer learning}

We next examine cross-site transfer learning, where a model trained on one or more source datasets is adapted to a target institution with limited labeled data.
This setting reflects practical deployment scenarios in which a pretrained model must be transferred to a new hospital without extensive feature engineering or large-scale retraining.

We focus on ReMed with LLM-based linguistic alignment and an English tokenizer, which performs best in pooled learning, and compare it against YAIB as a representative feature-aligned baseline.
For each ordered source--target dataset pair, we evaluate two adaptation strategies.
In the few-shot setting, models are fine-tuned using only 10\% of the target training data.
In the full fine-tuning setting, the entire target training dataset is used.

\begin{figure*}[ht] 
    \includegraphics[width=1\textwidth]{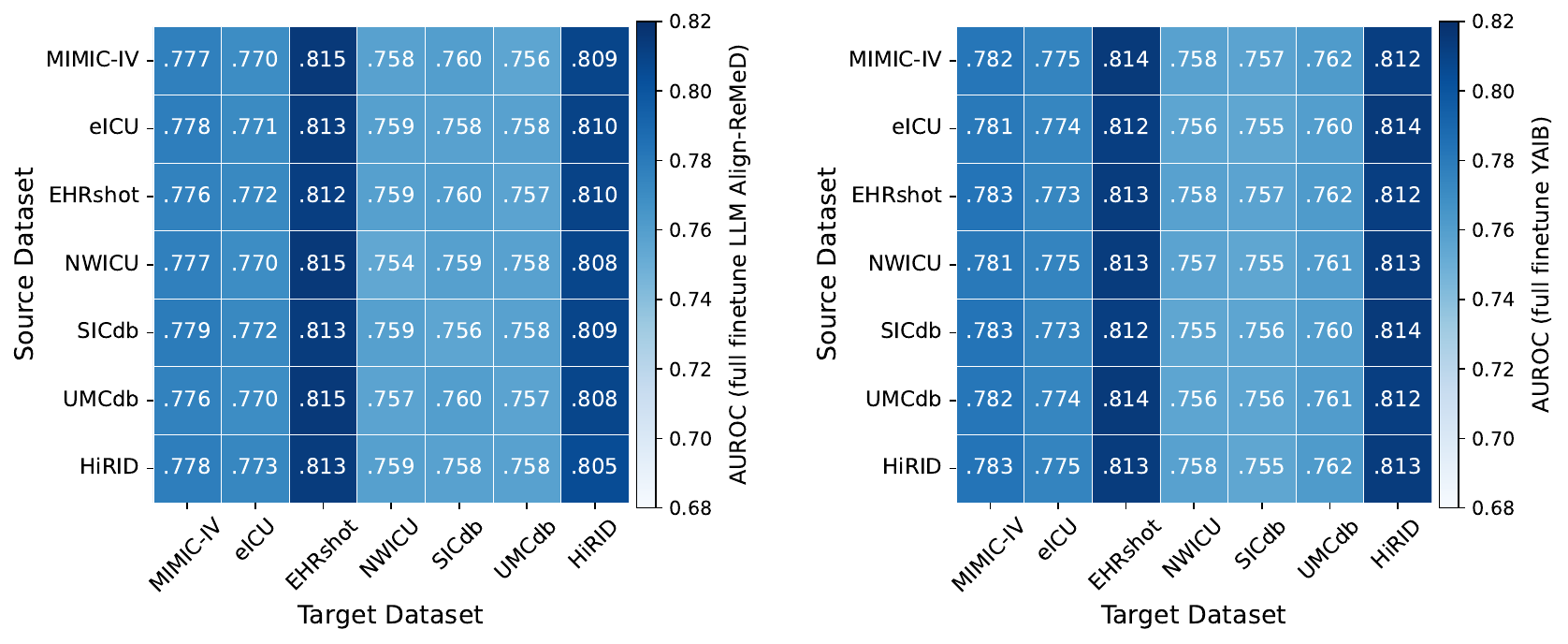}
    \caption{
   Cross-site transfer learning performance across seven datasets when fully finetuning of the target data for ReMed+LLM Align (left) and YAIB (right). Each heatmap shows AUROC when training on the source dataset (rows) and evaluating on the target dataset (columns); diagonal cells represents the single-dataset models (no transfer) , while off-diagonal cells correspond to transfer learning via fine-tuning on the target site. 
    }
    \label{fig:fig6}
    \vskip -4pt
\end{figure*}

Figure~\ref{fig:fig6} summarizes the results across all transfer directions.
In the few-shot setting, ReMed and YAIB achieve comparable average performance, with neither method consistently dominating across all source--target pairs.
In the full fine-tuning setting, both methods recover performance close to their respective single-dataset baselines.
These results indicate that schema-agnostic text-based models can be effectively adapted to new institutions with limited supervision, while retaining competitive performance when sufficient target data are available.

\subsection{Ablation study}

We conduct ablation studies to analyze the contribution of linguistic alignment and the choice of pretrained text encoder.

First, we examine different strategies for lingual alignment within ReMed.
We compare no explicit alignment with English tokenization (EnTok), dictionary-based translation (DictTrans), and LLM-based alignment using Medgemma-7b-it or Qwen3-Instruct-8b.
Experiments are conducted under both \textit{Single} and \textit{Multi} training regimes.
Under \textit{Single} training, all alignment variants yield similar performance, indicating that explicit alignment provides limited benefit when training and evaluation distributions are matched.
Under \textit{Multi} training, stronger alignment strategies lead to higher average AUROC, with Qwen3-Instruct-8b-based alignment achieving the best overall performance (Table~\ref{tab:ablation}).

Second, we evaluate sensitivity to the choice of pretrained text encoder under the best-performing alignment configuration.
We compare English encoders (GenHPF-SSL, BERT-mini, Bio-Clinical-BERT) and multilingual encoders (GenHPF-SSL, DistilBERT-multilingual, XLM-RoBERTa), all fine-tuned end-to-end under identical settings.
As shown in Table~\ref{tab:encoder}, performance differences across encoders are relatively small, suggesting that the observed gains are primarily driven by schema and language handling rather than the specific backbone architecture.

\begin{table}[t]

\centering
\renewcommand{\arraystretch}{1.5}
{\setlength{\tabcolsep}{4.5pt}
\newcommand{\mstd}[2]{#1 {\tiny (#2)}}
\begin{tabular}{llcccc}
\toprule
\multirow{2}{*}{Dataset} & \multirow{2}{*}{\shortstack{Train\\mode}}
& \multicolumn{1}{c}{No align}
& \multirow{2}{*}{\centering DictTrans}
& \multicolumn{2}{c}{LLM Align} \\
& & \multicolumn{1}{c}{EnTok} & & \multicolumn{1}{c}{MedGemma} & \multicolumn{1}{c}{Qwen3} \\
\cmidrule(lr){3-3}\cmidrule(lr){4-4}\cmidrule(lr){5-6}

\multirow{2}{*}{MIMIC-IV}
  & Single & \mstd{\underline{0.775}}{0.005} & \mstd{0.770}{0.005} & \mstd{0.773}{0.005} & \mstd{\textbf{0.777}}{0.006} \\
  & Multi  & \mstd{\underline{0.780}}{0.007} & \mstd{0.779}{0.008} & \mstd{\underline{0.780}}{0.008} & \mstd{\textbf{0.783}}{0.009} \\
\cmidrule(lr){1-6}

\multirow{2}{*}{eICU}
  & Single & \mstd{\underline{0.769}}{0.004} & \mstd{0.756}{0.004} & \mstd{\underline{0.769}}{0.005} & \mstd{\textbf{0.771}}{0.006} \\
  & Multi  & \mstd{0.772}{0.007} & \mstd{0.775}{0.008} & \mstd{\textbf{0.782}}{0.009} & \mstd{\underline{0.780}}{0.009} \\
\cmidrule(lr){1-6}

\multirow{2}{*}{EHRshot}
  & Single & \mstd{\textbf{0.812}}{0.005} & \mstd{0.804}{0.005} & \mstd{\underline{0.810}}{0.004} & \mstd{\textbf{0.812}}{0.006} \\
  & Multi  & \mstd{0.815}{0.008} & \mstd{\textbf{0.820}}{0.008} & \mstd{0.815}{0.008} & \mstd{\underline{0.817}}{0.009} \\
\cmidrule(lr){1-6}

\multirow{2}{*}{NWICU}
  & Single & \mstd{\textbf{0.756}}{0.005} & \mstd{0.752}{0.005} & \mstd{\underline{0.755}}{0.004} & \mstd{0.754}{0.006} \\
  & Multi  & \mstd{\underline{0.766}}{0.008} & \mstd{0.760}{0.009} & \mstd{0.763}{0.008} & \mstd{\textbf{0.770}}{0.010} \\
\cmidrule(lr){1-6}

\multirow{2}{*}{SICdb}
  & Single & \mstd{\underline{0.757}}{0.005} & \mstd{0.753}{0.005} & \mstd{\textbf{0.761}}{0.005} & \mstd{0.756}{0.006} \\
  & Multi  & \mstd{\underline{0.763}}{0.008} & \mstd{0.761}{0.009} & \mstd{0.760}{0.009} & \mstd{\textbf{0.765}}{0.010} \\
\cmidrule(lr){1-6}

\multirow{2}{*}{UMCdb}
  & Single & \mstd{0.755}{0.005} & \mstd{0.752}{0.005} & \mstd{\textbf{0.759}}{0.004} & \mstd{\underline{0.757}}{0.006} \\
  & Multi  & \mstd{0.757}{0.008} & \mstd{\underline{0.761}}{0.009} & \mstd{0.760}{0.009} & \mstd{\textbf{0.762}}{0.010} \\
\cmidrule(lr){1-6}

\multirow{2}{*}{HiRiD}
  & Single & \mstd{\underline{0.807}}{0.005} & \mstd{\textbf{0.808}}{0.005} & \mstd{0.805}{0.004} & \mstd{0.805}{0.006} \\
  & Multi  & \mstd{0.811}{0.008} & \mstd{\underline{0.812}}{0.009} & \mstd{\underline{0.812}}{0.009} & \mstd{\textbf{0.813}}{0.010} \\

\hline\hline
\multirow{2}{*}{Avg}
  & Single & \mstd{\textbf{0.776}}{0.004} & \mstd{\underline{0.771}}{0.004} & \mstd{\textbf{0.776}}{0.004} & \mstd{\textbf{0.776}}{0.005} \\
  & Multi  & \mstd{0.784}{0.005} & \mstd{0.785}{0.005} & \mstd{\underline{0.786}}{0.005} & \mstd{\textbf{0.788}}{0.006} \\
\bottomrule
\end{tabular}
}
\caption{\label{tab:ablation} Ablation study of lingual alignment strategies. We evaluate (i) no alignment and using English tokenization (No align; EnTok), (ii) dictionary-based translation (DictTrans), and (iii) LLM-based alignment using MedGemma and Qwen3.
Entries report AUROC averaged over all prediction tasks and five random seeds, with standard deviation in parentheses. Bold and underlined values denote the best and second-best performance within each dataset and training regime, respectively.}

\end{table}

\begin{table}[ht]
\centering
\renewcommand{\arraystretch}{1.3}

\begin{tabular}{lccc}
\toprule
\multicolumn{4}{c}{EnTok} \\
\midrule
        & GenHPF-SSL & BERT-mini & Bio-Clinical-BERT \\
Single  & 0.776 & 0.775 & 0.776 \\
Multi   & 0.788 & 0.787 & 0.788 \\
\midrule
\multicolumn{4}{c}{MLTok} \\
\midrule
        & GenHPF-SSL & DistilBERT-multi-lingual & XLM-RoBERTa \\
Single  & 0.773 & 0.772 & 0.772 \\
Multi   & 0.783 & 0.782 & 0.782 \\
\bottomrule
\end{tabular}

\caption{\label{tab:encoder} Comparison of English-only (EnTok) and multilingual (MLTok) text encoders within ReMed. AUROC values are averaged over all prediction tasks and five random seeds. GenHPF-SSL denotes the contrastive self-supervised encoder introduced in GenHPF. The EnTok block compares English encoders (BERT-mini, Bio-Clinical-BERT), while the MLTok block compares multilingual encoders (DistilBERT-multilingual, XLM-RoBERTa).}
\end{table}

\section{Discussion} \label{sec:discussion}

Our experiments reveal two consistent patterns. First, schema-agnostic text representations become particularly advantageous when learning is performed across heterogeneous institutions. Second, mapping non-English clinical tokens into a shared English space is a pragmatic and robust strategy for multilingual ICU EHR prediction in our setting, especially under pooled training.

A key observation is the shift between the \textit{Single} and \textit{Multi} regimes: YAIB is strongest when each site is modeled independently, whereas text-based models (GenHPF/ReMed) gain more reliably from multi-institutional learning. 
YAIB relies on a carefully curated shared feature set; under single-site training, this can serve as a strong regularizer that stabilizes learning in the presence of site-specific noise, label imbalance, and measurement idiosyncrasies. 
However, the same design can constrain what is shareable across sites: if overlap in engineered features is limited, the transferability that a pooled YAIB model can exploit may saturate quickly.
In contrast, schema-agnostic event-text linearization preserves a broader slice of each site’s raw event content and metadata.
When trained in the \textit{Multi} regime, this richer representation offers more opportunities to learn cross-site invariances (e.g., shared clinical semantics expressed through different local codes or table schemas) without requiring explicit feature harmonization.
Empirically, this is reflected by the larger and more consistent improvements of text-based models under pooled training compared with code-based baselines.

Across multilingual configurations, English-space unification via LLM alignment combined with an English tokenizer emerges as the most reliable option for pooled learning.
Multilingual tokenization (MLTok) underperforms EnTok/LLM Align on average, which is consistent with tokenization efficiency considerations: a generic multilingual vocabulary distributes capacity across many languages beyond those represented in our datasets, increasing subword fragmentation and sequence length for domain-specific clinical terms.
Longer, more fragmented sequences effectively enlarge the optimization search space for the downstream encoder under fixed compute budgets, potentially making it harder to learn stable cross-site representations.
Translation-based unification sidesteps this issue by mapping Dutch/German tokens into a shared English space while keeping the rest of the pipeline unchanged, enabling the model to leverage strong English tokenization and established English clinical encoders.
For predominantly English datasets, EnTok and LLM Align often coincide in the \textit{Single} setting because translation is rarely triggered, which explains the frequent ties between these configurations under single-site training.

Transfer learning results provide practical guidance for deployment. In the 10\% few-shot setting, ReMed (LLM Align) achieves broadly comparable performance to YAIB across many source--target directions, but does so without any dataset-specific feature engineering or cross-dataset harmonization. This is operationally meaningful: when onboarding a new hospital with limited labeled data, reducing the need for manual feature alignment can significantly lower time-to-model and engineering overhead. When abundant labeled data are available at the target site, both YAIB and ReMed recover performance near their single-site baselines after full fine-tuning, indicating that the main advantage of schema-agnostic text modeling is not necessarily asymptotic accuracy, but rather scalability and adaptability under heterogeneity and limited target supervision.

The ablation results indicate that linguistic alignment matters primarily under pooled multi-institutional training: in the \textit{Single} regime, explicit word-level alignment provides little benefit because the model can accommodate site- and language-specific artifacts when train/test distributions match, whereas in the \textit{Multi} regime alignment becomes beneficial by reducing avoidable linguistic variance and allowing the encoder to focus on shared clinical semantics across languages and documentation styles.
Dictionary-based translation can underperform due to limited coverage of abbreviations and inflections that leaves mixed-language inputs, while LLM-based alignment yields more consistent gains by resolving more tokens contextually; Qwen3-Instruct-8b slightly but consistently outperforms Medgemma-7b-it, suggesting robustness is driven not only by biomedical tuning but also general instruction-following and contextual translation quality.
In contrast, varying the pretrained backbone produces only small differences after end-to-end fine-tuning, implying that future improvements are more likely to come from better heterogeneity handling (schema/language), more reliable alignment for rare or ambiguous terms, and training objectives that discourage institution-specific shortcuts; adding confidence-aware alignment or targeted lexicon refinement may further improve robustness.

\section{Limitation}
Although our work demonstrates the feasibility of multilingual, multi-institutional EHR prediction using text-based models, several limitations still remain. 
First, although we aggregate seven publicly available ICU datasets from multiple countries, the language coverage is still limited to a small set of languages.
Second, our approach relies on off-the-shelf LLM-based word-level alignment and does not explicitly account for translation or alignment errors, which may affect representations of rare clinical terms.
Finally, our study focuses on lightweight word-level alignment strategies and does not explore more computationally intensive context-aware translation methods (e.g. using LLM as a text-encoder), which may further improve prediction performance.

\section{Conclusion}
In this study, we empirically investigated multilingual, multi-institutional EHR prediction using text-based models across heterogeneous datasets.
By converting structured EHR events into text and applying lightweight LLM-based word-level alignment, we examined how existing text-based models perform when trained on pooled multilingual data.
Across seven public ICU datasets, translation-based text models consistently outperform baselines based on domain-driven feature selection or monolingual modeling, and show strong performance in pooled training and cross-language transfer.
These results suggest that simple language alignment combined with schema-agnostic text representations offers a practical alternative to costly common data model transformations for multilingual, multi-institutional EHR prediction.
Future work will extend this analysis to broader language coverage, additional clinical tasks, and multimodal settings, and examine fairness- and privacy-aware training.

\section{Data and Code Availability}
All datasets used in this work are publicly available.
Our code implementation is available on Github at \url{https://github.com/hoon9405/Multi-lingual-EHR-prediction}.


\begin{table*}[t]
\centering
\caption{\label{tab:multi-class} Statistics for multi-class classification tasks. All numbers are represented as percent.}
\renewcommand{\arraystretch}{1.05}
{\setlength{\tabcolsep}{5pt}
\scriptsize
\begin{tabular}{lrrrrrr rrrrrr rrrrrr}
\toprule
\multirow{2}{*}{Label}
  & \multicolumn{6}{c}{MIMIC-IV}
  & \multicolumn{6}{c}{eICU}
  & \multicolumn{6}{c}{EHRshot} \\
\cmidrule(lr){2-7}\cmidrule(lr){8-13}\cmidrule(lr){14-19}
  & -1 & 0 & 1 & 2 & 3 & 4
  & -1 & 0 & 1 & 2 & 3 & 4
  & -1 & 0 & 1 & 2 & 3 & 4 \\
\midrule
creatinine\_1 & 15.15 & 60.20 & 14.36 &  6.67 &  2.24 &  1.37 & 20.74 & 51.77 & 15.16 &  7.51 &  2.59 &  2.24 & 19.01 & 46.88 & 19.18 &  9.21 &  3.41 &  2.31 \\
creatinine\_2 & 22.73 & 56.22 & 12.51 &  5.55 &  1.83 &  1.16 & 32.50 & 45.06 & 12.35 &  6.03 &  2.14 &  1.91 & 25.15 & 43.88 & 17.70 &  8.42 &  2.48 &  2.38 \\
creatinine\_3 & 32.58 & 49.13 & 11.06 &  4.74 &  1.58 &  0.92 & 43.33 & 38.02 & 10.27 &  5.05 &  1.77 &  1.58 & 30.94 & 40.22 & 16.80 &  7.52 &  2.79 &  1.72 \\
platelets\_1  &  8.20 & 54.78 & 20.61 & 12.39 &  3.43 &  0.60 & 20.27 & 47.51 & 18.35 & 10.64 &  2.65 &  0.57 & 20.94 & 38.22 & 17.97 & 15.80 &  5.04 &  2.04 \\
platelets\_2  & 15.93 & 54.60 & 16.08 &  9.57 &  3.24 &  0.58 & 32.55 & 42.12 & 13.99 &  8.38 &  2.42 &  0.55 & 26.91 & 38.91 & 15.21 & 12.49 &  4.38 &  2.10 \\
platelets\_3  & 26.41 & 50.64 & 12.08 &  7.43 &  2.92 &  0.53 & 43.74 & 37.07 & 10.28 &  6.34 &  2.06 &  0.51 & 32.80 & 38.91 & 11.69 & 10.52 &  4.31 &  1.76 \\
urine\_1      & 16.07 & 74.49 &  6.25 &  2.20 &  0.38 &  0.62 & 42.02 & 51.30 &  2.25 &  1.03 &  0.21 &  3.19 & 18.08 & 70.99 &  3.86 &  2.00 &  0.76 &  4.31 \\
urine\_2      & 46.39 & 47.98 &  3.70 &  1.34 &  0.26 &  0.34 & 54.85 & 40.30 &  1.42 &  0.69 &  0.15 &  2.60 & 24.97 & 66.47 &  2.73 &  1.72 &  0.48 &  3.62 \\
urine\_3      & 62.55 & 33.69 &  2.47 &  0.94 &  0.11 &  0.24 & 63.79 & 32.50 &  1.00 &  0.46 &  0.14 &  2.12 & 31.29 & 61.50 &  1.66 &  1.97 &  0.38 &  3.21 \\
AKI\_1        & 21.16 & 57.55 & 17.08 &  2.42 &  0.86 &  0.92 & 42.02 & 45.74 &  7.09 &  1.27 &  0.45 &  3.43 & --    & --    & --    & --    & --    & --    \\
AKI\_2        & 49.64 & 36.15 & 11.28 &  1.54 &  0.71 &  0.68 & 54.85 & 35.97 &  5.13 &  0.87 &  0.36 &  2.82 & --    & --    & --    & --    & --    & --    \\
AKI\_3        & 64.86 & 24.88 &  8.05 &  1.11 &  0.51 &  0.60 & 63.79 & 28.84 &  4.11 &  0.61 &  0.31 &  2.34 & --    & --    & --    & --    & --    & --    \\
hb\_1         &  8.21 & 14.22 & 40.14 & 25.30 & 12.13 &  0.00 & 18.89 & 10.91 & 32.74 & 23.50 & 13.96 &  0.00 & 20.83 & 17.28 & 38.39 & 16.07 &  7.42 &  0.00 \\
hb\_2         & 15.95 & 12.66 & 37.86 & 23.23 & 10.30 &  0.00 & 31.28 &  8.72 & 29.04 & 19.99 & 10.97 &  0.00 & 26.80 & 15.59 & 36.67 & 14.76 &  6.17 &  0.00 \\
hb\_3         & 26.40 & 11.51 & 33.52 & 19.87 &  8.70 &  0.00 & 42.67 &  7.12 & 24.86 & 16.51 &  8.84 &  0.00 & 32.84 & 14.00 & 34.77 & 13.21 &  5.17 &  0.00 \\
wbc\_1        &  8.20 &  4.10 & 59.17 & 28.53 &  0.00 &  0.00 & 19.79 &  2.71 & 50.77 & 26.73 &  0.00 &  0.00 & 20.83 &  6.24 & 45.33 & 27.60 &  0.00 &  0.00 \\
wbc\_2        & 15.89 &  4.43 & 57.92 & 21.76 &  0.00 &  0.00 & 32.13 &  2.45 & 45.49 & 19.93 &  0.00 &  0.00 & 26.80 &  6.11 & 44.64 & 22.46 &  0.00 &  0.00 \\
wbc\_3        & 26.32 &  4.04 & 51.21 & 18.43 &  0.00 &  0.00 & 43.35 &  2.02 & 37.81 & 16.82 &  0.00 &  0.00 & 32.84 &  5.28 & 42.39 & 19.49 &  0.00 &  0.00 \\
bicarbonate\_1&  7.53 & 25.07 & 53.58 & 13.81 &  0.00 &  0.00 & 20.95 & 15.86 & 46.21 & 16.98 &  0.00 &  0.00 & 19.04 & 11.21 & 48.67 & 21.08 &  0.00 &  0.00 \\
bicarbonate\_2& 15.52 & 19.10 & 49.11 & 16.27 &  0.00 &  0.00 & 32.87 & 11.06 & 38.77 & 17.30 &  0.00 &  0.00 & 25.22 &  9.59 & 43.26 & 21.94 &  0.00 &  0.00 \\
bicarbonate\_3& 25.85 & 14.91 & 43.14 & 16.09 &  0.00 &  0.00 & 43.56 &  7.94 & 31.71 & 16.78 &  0.00 &  0.00 & 30.98 &  8.00 & 40.67 & 20.35 &  0.00 &  0.00 \\
sodium\_1     &  7.33 & 21.86 & 64.38 &  6.43 &  0.00 &  0.00 & 16.12 & 14.32 & 59.13 & 10.42 &  0.00 &  0.00 & 18.63 & 20.66 & 52.50 &  8.21 &  0.00 &  0.00 \\
sodium\_2     & 14.77 & 17.74 & 60.57 &  6.93 &  0.00 &  0.00 & 28.46 & 11.54 & 50.31 &  9.68 &  0.00 &  0.00 & 24.91 & 19.25 & 48.05 &  7.80 &  0.00 &  0.00 \\
sodium\_3     & 24.91 & 14.90 & 53.33 &  6.86 &  0.00 &  0.00 & 39.66 &  9.40 & 42.20 &  8.74 &  0.00 &  0.00 & 30.80 & 18.80 & 43.05 &  7.35 &  0.00 &  0.00 \\
\bottomrule
\end{tabular}
}

{\scriptsize
\begin{tabular}{lrrrrrr rrrrrr rrrrrr}
\toprule
\multirow{2}{*}{Label}
  & \multicolumn{6}{c}{nwicu}
  & \multicolumn{6}{c}{UMCdb}
  & \multicolumn{6}{c}{HiRiD} \\
\cmidrule(lr){2-7}\cmidrule(lr){8-13}\cmidrule(lr){14-19}
  & -1 & 0 & 1 & 2 & 3 & 4
  & -1 & 0 & 1 & 2 & 3 & 4
  & -1 & 0 & 1 & 2 & 3 & 4 \\
\midrule
creatinine\_1 &  9.87 & 57.38 & 16.44 &  9.74 &  3.47 &  3.10 & 37.95 & 48.07 &  8.49 &  4.08 &  1.08 &  0.33 & 35.03 & 47.17 & 10.03 &  5.59 &  1.61 &  0.58 \\
creatinine\_2 & 18.01 & 53.48 & 14.47 &  8.12 &  3.05 &  2.86 & 53.70 & 36.63 &  5.92 &  2.76 &  0.72 &  0.26 & 56.60 & 32.64 &  6.47 &  3.04 &  0.85 &  0.40 \\
creatinine\_3 & 26.95 & 48.02 & 12.75 &  6.94 &  2.74 &  2.60 & 62.75 & 29.87 &  4.81 &  1.84 &  0.50 &  0.23 & 69.63 & 23.59 &  4.43 &  1.66 &  0.45 &  0.23 \\
platelets\_1  & 12.31 & 55.99 & 17.76 & 10.01 &  3.02 &  0.90 & 11.87 & 48.58 & 20.12 & 13.59 &  4.39 &  1.45 & 22.79 & 39.77 & 16.81 & 14.27 &  4.70 &  1.67 \\
platelets\_2  & 20.32 & 52.54 & 15.00 &  8.37 &  2.95 &  0.81 & 29.78 & 40.00 & 14.52 & 10.32 &  3.96 &  1.41 & 45.87 & 29.52 & 10.73 &  8.49 &  3.91 &  1.49 \\
platelets\_3  & 29.19 & 48.99 & 11.18 &  7.15 &  2.74 &  0.76 & 40.59 & 36.45 & 10.55 &  7.68 &  3.58 &  1.15 & 59.92 & 24.23 &  6.68 &  4.97 &  2.98 &  1.21 \\
urine\_1      &100.00 &  0.00 &  0.00 &  0.00 &  0.00 &  0.00 & 31.86 & 63.90 &  2.57 &  1.14 &  0.19 &  0.35 & 21.08 & 69.63 &  5.97 &  2.00 &  0.47 &  0.84 \\
urine\_2      &100.00 &  0.00 &  0.00 &  0.00 &  0.00 &  0.00 & 50.16 & 48.08 &  1.17 &  0.35 &  0.08 &  0.16 & 46.58 & 49.10 &  2.75 &  1.04 &  0.22 &  0.32 \\
urine\_3      &100.00 &  0.00 &  0.00 &  0.00 &  0.00 &  0.00 & 60.28 & 38.95 &  0.49 &  0.23 &  0.01 &  0.04 & 61.10 & 36.77 &  1.32 &  0.57 &  0.08 &  0.16 \\
AKI\_1        &  9.87 & 90.00 &  0.12 &  0.01 &  0.00 &  0.00 & 37.95 & 51.80 &  8.61 &  0.74 &  0.54 &  0.36 & 18.96 & 66.65 & 10.70 &  2.12 &  0.67 &  0.90 \\
AKI\_2        & 18.01 & 78.52 &  3.33 &  0.03 &  0.04 &  0.07 & 53.70 & 39.42 &  5.61 &  0.53 &  0.41 &  0.33 & 45.31 & 46.71 &  6.08 &  1.05 &  0.46 &  0.39 \\
AKI\_3        & 26.95 & 68.27 &  4.46 &  0.13 &  0.08 &  0.10 & 62.75 & 31.79 &  4.59 &  0.35 &  0.25 &  0.28 & 60.50 & 34.87 &  3.54 &  0.66 &  0.20 &  0.24 \\
hb\_1         & 14.65 & 14.89 & 29.07 & 21.28 & 20.11 &  0.00 &  8.63 &  5.07 & 39.74 & 32.56 & 14.00 &  0.00 & 21.90 &  5.73 & 39.79 & 23.26 &  9.32 &  0.00 \\
hb\_2         & 22.54 & 12.85 & 27.94 & 19.43 & 17.23 &  0.00 & 27.93 &  4.14 & 32.65 & 25.94 &  9.34 &  0.00 & 45.19 &  3.45 & 29.04 & 16.46 &  5.85 &  0.00 \\
hb\_3         & 31.24 & 10.98 & 26.53 & 16.79 & 14.46 &  0.00 & 39.16 &  3.30 & 27.80 & 22.08 &  7.66 &  0.00 & 59.18 &  2.50 & 21.89 & 12.23 &  4.19 &  0.00 \\
wbc\_1        & 12.26 &  4.05 & 55.88 & 27.81 &  0.00 &  0.00 & 11.86 &  2.58 & 42.01 & 43.55 &  0.00 &  0.00 & 21.78 &  2.83 & 50.26 & 25.13 &  0.00 &  0.00 \\
wbc\_2        & 20.26 &  3.89 & 53.62 & 22.22 &  0.00 &  0.00 & 29.79 &  2.03 & 35.90 & 32.28 &  0.00 &  0.00 & 45.18 &  2.00 & 36.38 & 16.44 &  0.00 &  0.00 \\
wbc\_3        & 29.20 &  3.35 & 48.02 & 19.43 &  0.00 &  0.00 & 40.58 &  1.67 & 31.34 & 26.40 &  0.00 &  0.00 & 59.13 &  1.47 & 26.01 & 13.39 &  0.00 &  0.00 \\
bicarbonate\_1& 10.54 & 18.14 & 55.36 & 15.96 &  0.00 &  0.00 & 11.26 & 23.47 & 53.08 & 12.19 &  0.00 &  0.00 & 27.62 & 10.25 & 52.92 &  9.21 &  0.00 &  0.00 \\
bicarbonate\_2& 18.52 & 14.03 & 49.89 & 17.56 &  0.00 &  0.00 & 29.61 & 12.67 & 42.77 & 14.95 &  0.00 &  0.00 & 49.83 &  5.11 & 36.35 &  8.71 &  0.00 &  0.00 \\
bicarbonate\_3& 27.46 & 11.11 & 43.82 & 17.61 &  0.00 &  0.00 & 40.42 &  7.26 & 35.26 & 17.06 &  0.00 &  0.00 & 62.93 &  3.58 & 26.05 &  7.44 &  0.00 &  0.00 \\
sodium\_1     & 10.41 & 15.91 & 64.33 &  9.34 &  0.00 &  0.00 &  8.36 & 11.23 & 66.66 & 13.75 &  0.00 &  0.00 & 17.60 & 15.35 & 60.86 &  6.19 &  0.00 &  0.00 \\
sodium\_2     & 18.38 & 13.54 & 58.24 &  9.83 &  0.00 &  0.00 & 27.73 &  7.80 & 49.99 & 14.48 &  0.00 &  0.00 & 42.05 &  9.80 & 42.36 &  5.79 &  0.00 &  0.00 \\
sodium\_3     & 27.31 & 11.44 & 51.46 &  9.79 &  0.00 &  0.00 & 38.96 &  6.56 & 39.91 & 14.57 &  0.00 &  0.00 & 56.84 &  7.13 & 30.55 &  5.47 &  0.00 &  0.00 \\
\bottomrule
\end{tabular}
}
{\setlength{\tabcolsep}{5pt}
\scriptsize
\hspace*{-3em} 
\begin{minipage}[t]{0.48\textwidth}
\centering
\begin{tabular}{lrrrrrr}
Label & \multicolumn{6}{c}{HiRID} \\
\midrule
creatinine\_1 & 25.69 & 55.80 & 11.66 &  4.92 &  1.16 &  0.75 \\
creatinine\_2 & 44.61 & 41.32 &  8.77 &  3.90 &  0.81 &  0.59 \\
creatinine\_3 & 56.44 & 32.86 &  6.83 &  2.87 &  0.59 &  0.41 \\
platelets\_1  & 21.02 & 42.61 & 22.91 & 11.13 &  1.97 &  0.36 \\
platelets\_2  & 39.31 & 36.00 & 15.19 &  7.41 &  1.75 &  0.34 \\
platelets\_3  & 51.25 & 33.13 &  9.18 &  4.64 &  1.51 &  0.29 \\
urine\_1      & 24.72 & 54.59 & 14.18 &  4.78 &  0.40 &  1.34 \\
urine\_2      & 48.57 & 40.40 &  7.47 &  2.42 &  0.22 &  0.92 \\
urine\_3      & 63.21 & 30.02 &  4.70 &  1.35 &  0.09 &  0.63 \\
AKI\_1        & 33.48 & 48.08 & 12.91 &  4.00 &  0.31 &  1.22 \\
AKI\_2        & 54.20 & 34.97 &  7.72 &  2.06 &  0.20 &  0.85 \\
AKI\_3        & 67.05 & 25.25 &  5.81 &  1.13 &  0.12 &  0.63 \\
hb\_1         &  9.31 & 16.01 & 42.90 & 22.46 &  9.33 &  0.00 \\
hb\_2         & 28.60 & 13.14 & 35.99 & 15.77 &  6.49 &  0.00 \\
hb\_3         & 41.82 & 10.05 & 30.68 & 12.41 &  5.05 &  0.00 \\
wbc\_1        & 21.01 &  1.95 & 54.86 & 22.18 &  0.00 &  0.00 \\
wbc\_2        & 39.33 &  1.94 & 43.54 & 15.20 &  0.00 &  0.00 \\
wbc\_3        & 51.24 &  1.40 & 35.09 & 12.28 &  0.00 &  0.00 \\
bicarbonate\_1&  9.25 &  9.94 & 65.33 & 15.47 &  0.00 &  0.00 \\
bicarbonate\_2& 28.52 &  6.42 & 48.08 & 16.98 &  0.00 &  0.00 \\
bicarbonate\_3& 41.65 &  4.95 & 38.55 & 14.84 &  0.00 &  0.00 \\
sodium\_1     & 34.89 & 10.35 & 48.11 &  6.65 &  0.00 &  0.00 \\
sodium\_2     & 49.03 &  7.26 & 36.81 &  6.90 &  0.00 &  0.00 \\
sodium\_3     & 57.65 &  5.39 & 30.50 &  6.47 &  0.00 &  0.00 \\
\bottomrule
\end{tabular}
\end{minipage}
}
\hfill
\begin{minipage}[t]{0.48\textwidth}
\vspace*{-12\baselineskip} 
\centering
\caption{\label{tab:binary-class} Statistics for binary-class classification tasks (\%).}
\begin{tabular}{l*{3}{cc}}
\toprule
\multicolumn{7}{c}{LOS and mortality (MIMIC-IV, eICU, EHRshot)} \\
\midrule
& \multicolumn{2}{c}{MIMIC-IV}
& \multicolumn{2}{c}{eICU}
& \multicolumn{2}{c}{EHRshot} \\
\cmidrule(lr){2-3}\cmidrule(lr){4-5}\cmidrule(lr){6-7}
Dataset & 0 & 1 & 0 & 1 & 0 & 1 \\
\midrule
los\_7        & 79.78 & 20.22 & 80.76 & 19.24 & 49.39 & 50.61 \\
los\_14       & 93.63 &  6.37 & 94.58 &  5.42 & 69.59 & 30.41 \\
mortality\_1  & 97.79 &  2.21 & 98.37 &  1.63 & 40.31 & 59.69 \\
mortality\_2  & 96.42 &  3.58 & 96.94 &  3.06 & 32.73 & 67.27 \\
mortality\_3  & 95.19 &  4.81 & 95.75 &  4.25 & 29.96 & 70.04 \\
mortality\_7  & 91.86 &  8.14 & 92.79 &  7.21 & 24.97 & 75.03 \\
mortality\_14 & 88.89 & 11.11 & 90.39 &  9.61 & 22.86 & 77.14 \\
\bottomrule
\end{tabular}

\vspace{0.6em}

\begin{tabular}{l*{3}{cc}}
\toprule
\multicolumn{7}{c}{LOS and mortality (NWICU, UMCdb, SICdb)} \\
\midrule
& \multicolumn{2}{c}{NWICU}
& \multicolumn{2}{c}{UMCdb}
& \multicolumn{2}{c}{SICdb} \\
\cmidrule(lr){2-3}\cmidrule(lr){4-5}\cmidrule(lr){6-7}
Dataset & 0 & 1 & 0 & 1 & 0 & 1 \\
\midrule
los\_7        & 75.16 & 24.84 & 56.29 & 43.71 & 74.94 & 25.06 \\
los\_14       & 90.65 &  9.35 & 76.70 & 23.30 & 89.52 & 10.48 \\
mortality\_1  & 97.89 &  2.11 & 96.73 &  3.27 & 98.68 &  1.32 \\
mortality\_2  & 96.55 &  3.45 & 94.46 &  5.54 & 97.78 &  2.22 \\
mortality\_3  & 95.37 &  4.63 & 93.13 &  6.87 & 97.10 &  2.90 \\
mortality\_7  & 91.55 &  8.45 & 89.84 & 10.16 & 94.83 &  5.17 \\
mortality\_14 & 87.47 & 12.53 & 86.74 & 13.26 & 92.14 &  7.86 \\
\bottomrule
\end{tabular}
\end{minipage}

\end{table*}

\section{Appendix}
\label{sec:appendix}
\subsection{Statistics for prediction tasks}\label{apd:data}
This section summarizes the label distributions for all prediction tasks.
All values in the tables denote the proportion of ICU stays (in \%) that fall into each class for every dataset.

\noindent\textbf{Binary classification tasks (Table~\ref{tab:binary-class}).}
We report class ratios for all binary tasks used in the main experiments:
$k$-day mortality for $k\in\{1,2,3,7,14\}$ and $k$-day length of stay (LOS) for $k\in\{7,14\}$.
For each task, class ``0'' corresponds to the negative outcome (e.g., survival beyond the prediction horizon or LOS $\le k$ days), and class ``1'' corresponds to the positive outcome (e.g., death within $k$ days or LOS $> k$ days).
These statistics illustrate the varying degrees of class imbalance across institutions and time horizons.

\noindent\textbf{Multi-class classification tasks (Table~\ref{tab:multi-class}).}
We also summarize label compositions for the multi-class tasks, which include future lab severity for creatinine, platelets, white blood cell count (WBC), hemoglobin, bicarbonate, sodium, and urine output, as well as $k$-day AKI staging for $k\in\{1,2,3\}$.
For each horizon $k$, rows such as \texttt{creatinine\_k}, \texttt{platelets\_k}, or \texttt{AKI\_k} denote the corresponding prediction task.
Columns indexed by $0$--$4$ represent discrete severity or stage categories derived from clinically defined thresholds following YAIB, while the column ``$-1$'' denotes the \emph{Null} class.
For laboratory tasks, the Null class primarily corresponds to ICU stays affected by dialysis or missing measurements, and these samples are excluded from the loss for that task.
For AKI, classes $0$--$3$ follow KDIGO-like staging rules, with the Null class again indicating stays where staging cannot be determined.
Together, these tables provide an overview of label imbalance patterns that models must handle in our multilingual, multi-institutional setting.
Following YAIB, we discretize continuous lab values into ordinal bins (e.g., platelets and urine output into 5 bins; creatinine into 5 bins with cutoffs $\{1.2, 2.0, 3.5, 5.0\}$; WBC into 3 bins; hemoglobin into 4 bins; bicarbonate and sodium into 3 bins).
For AKI, we compute stage from relative creatinine change with respect to a pre-ICU baseline and, when urine output is available, define the final stage as the maximum of creatinine- and urine-based stages (NWICU uses creatinine only).

\subsection{Hyperparameters}
We explored various hyperparameters to determine the optimal for each framework.
However, we found that the impact of these hyperparameters on the results was not significant.
Consequently, we use a unified set of hyperparameters for all cases, thereby simplifying the experiment while maintaining the performance for each model.
The final hyperparameters are a dropout of 0.3, a batch size of 32, and a learning rate of 1e-4.
For models other than Rajkomar, we followed the original works in terms of the number of layers and the embedding dimensionality.
Rajkomar uses the same model architecture as GenHPF, but replaces text-based embeddings with code-based embeddings.
For pretraining GenHPF-SSL, we follow same hyperparameters provided at the original paper.

\subsection{Lingual alignment prompts}
\label{apd:prompt}

For word-level lingual alignment, we employ LLM-based translation only for tokens whose language cannot be resolved by lexicon-based matching.
Each unresolved token is translated independently using a short context-aware prompt, where the input consists of the original token and its surrounding event text (feature name and local context).

Specifically, we prompt the LLM as follows:
\begin{quote}
\small
\textit{``You are translating a clinical term from ICU electronic health records.
Given the following token and its context, translate the token into a concise English medical term.
Return only the translated term without explanation.
If the token is already English or the translation is uncertain, return the original token.''}
\end{quote}

The model output is accepted only when it is a single, unambiguous English term; otherwise, the original token is retained.
This conservative design prevents over-translation and preserves clinical abbreviations and proper nouns.
All translated tokens are then used as input to the downstream text encoder for multilingual alignment.

\subsection{Alignment of YAIB features across datasets}
\label{apd:yaib_align}

YAIB defines a fixed set of 48 time-series clinical variables covering vital signs, laboratory measurements, and interventions.
For datasets not originally included in YAIB (SICdb, NWICU, and EHRshot), we manually aligned raw EHR variables to the YAIB feature set to enable consistent model inputs across all datasets.

Specifically, for each YAIB variable, we identified semantically equivalent measurements in the target dataset based on variable name, clinical meaning, and unit consistency.
When multiple candidate variables existed, we selected the one most commonly measured in routine ICU care.
Variables without a reliable counterpart were treated as missing.
All aligned variables were resampled to the same temporal resolution and normalized using dataset-specific statistics following the original YAIB preprocessing pipeline.

This alignment procedure ensures that all seven datasets share an identical 48-dimensional input space for YAIB-based baselines, allowing fair comparison across institutions without introducing dataset-specific features.


\section*{REFERENCES}

\begin{thebibliography}{00}
\bibitem{choi2016doctor} 
E. Choi, M. T. Bahadori, A. Schuetz, W. F. Stewart, and J. Sun, 
``Doctor AI: Predicting clinical events via recurrent neural networks,'' 
in {\it Proc. Machine Learning for Healthcare}, pp. 301--318, 2016.

\bibitem{awad2017early} 
A. Awad, M. Bader-El-Den, J. McNicholas, and J. Briggs, 
``Early hospital mortality prediction of intensive care unit patients using an ensemble learning approach,'' 
{\it Int. J. Med. Inform.}, vol. 108, pp. 185--195, 2017.

\bibitem{thiel2010early} 
S. W. Thiel, J. M. Rosini, W. Shannon, J. A. Doherty, S. T. Micek, and M. H. Kollef, 
``Early prediction of septic shock in hospitalized patients,'' 
{\it J. Hosp. Med.}, vol. 5, no. 1, pp. 19--25, 2010.

\bibitem{shameer2017predictive} 
K. Shameer, K. W. Johnson, A. Yahi, R. Miotto, L. Li, D. Ricks, J. Jebakaran, P. Kovatch, 
P. Sengupta, S. Gelijns, {\it et al.}, 
``Predictive modeling of hospital readmission rates using electronic medical record-wide machine learning: 
a case-study using a Mount Sinai heart failure cohort,'' 
in {\it Proc. Pacific Symp. Biocomput.}, pp. 276--287, 2017.

\bibitem{hur2024genhpf}
K. Hur, J. Oh, J. Kim, J. Kim, M. J. Lee, E. Cho, S.-E. Moon, Y.-H. Kim, L. Atallah, and E. Choi,
``GenHPF: General healthcare predictive framework for multi-task multi-source learning,''
{\it IEEE Journal of Biomedical and Health Informatics}, vol. 28, no. 1, pp. 502--513, 2024.


\bibitem{rajkomar2018scalable}
A. Rajkomar, E. Oren, K. Chen, {\it et al.}, 
``Scalable and accurate deep learning with electronic health records,'' 
{\it npj Digital Medicine}, vol. 1, no. 1, p. 18, 2018.

\bibitem{meystre2017secondaryuse}
S. M. Meystre, G. Lovis, T. Bürkle, H. U. Tognola, D. Budrionis, and C. Lehmann, 
``Clinical data reuse or secondary use: Current status and potential future progress,'' 
{\it Yearbook of Med. Inform.}, vol. 26, no. 1, pp. 38--52, 2017.

\bibitem{hyland2020circfailure}
S. L. Hyland, M. Faltys, M. Hüser, X. Lyu, T. Gumbsch, C. Esteban, C. Bock, M. Horn, 
M. Moor, B. Rieck, {\it et al.}, 
``Early prediction of circulatory failure in the intensive care unit using machine learning,'' 
{\it Nat. Med.}, vol. 26, no. 3, pp. 364--373, 2020.

\bibitem{thoral2021amsterdamumcdb_ml}
P. J. Thoral, J. M. Peppink, R. H. Driessen, E. J. G. Sijbrands, E. J. O. Kompanje, 
L. Kaplan, H. Bailey, {\it et al.}, 
``Explainable machine learning on AmsterdamUMCdb for ICU discharge decision support,'' 
{\it Crit. Care}, vol. 25, no. 1, p. 304, 2021.

\bibitem{sauer2022publicicu}
C. M. Sauer, H. R. Sasson, M. C. Celi, and L. A. Celi, 
``Systematic review and comparison of publicly available intensive care unit databases,'' 
{\it Crit. Care Med.}, vol. 50, no. 8, pp. e685--e697, 2022.

\bibitem{hripcsak2015ohdsi}
G. Hripcsak, J. D. Duke, N. H. Shah, C. G. Reich, V. Huser, M. J. Schuemie, M. A. Suchard, 
R. W. Park, I. C. K. Wong, P. R. Rijnbeek, {\it et al.}, 
``Observational Health Data Sciences and Informatics (OHDSI): Opportunities for observational researchers,'' 
in {\it MEDINFO 2015: eHealth-Enabled Health}, 
Studies in Health Technology and Informatics, vol. 216, pp. 574--578, 2015.

\bibitem{park2019ohdsi}
R. W. Park, 
``Common data model and distributed research network: Observational Health Data Sciences and Informatics (OHDSI),'' 
{\it Korean J. Med.}, vol. 94, no. 4, pp. 309--314, 2019.

\bibitem{belenkaya2021extending}
R. Belenkaya, K. A. Chamberlin, T. D. Xie, {\it et al.}, 
``Extending the OMOP common data model and standardized vocabularies to support observational research in oncology,'' 
{\it J. Biomed. Inform.}, vol. 118, p. 103795, 2021.

\bibitem{yang2022gatortron}
X. Yang, A. Chen, N. PourNejatian, H.-C. Shin, K. E. Smith, C. Parisien, C. Compas, 
C. Martin, M. G. Flores, Y. Zhang, {\it et al.}, 
``GatorTron: A large clinical language model to unlock patient information from unstructured electronic health records,'' 
{\it npj Digital Medicine}, vol. 5, no. 1, p. 41, 2022.

\bibitem{singhal2023llmclinic}
K. Singhal, S. Azizi, T. Tu, S. S. Mahdavi, J. Wei, H. W. Chung, N. Scales, A. Tanwani, 
H. Cole-Lewis, S. Pfohl, {\it et al.}, 
``Large language models encode clinical knowledge,'' 
{\it Nature}, vol. 620, no. 7972, pp. 172--180, 2023.

\bibitem{johnson2016mimic3}
A. E. W. Johnson, T. J. Pollard, L. Shen, H. L. H. Lehman, M. Feng, 
M. Ghassemi, B. Moody, P. Szolovits, L. A. Celi, and R. G. Mark, 
``MIMIC-III, a freely accessible critical care database,'' 
{\it Sci. Data}, vol. 3, p. 160035, 2016.

\bibitem{pollard2018eicu}
T. J. Pollard, A. E. W. Johnson, J. Raffa, L. A. Celi, R. G. Mark, and O. Badawi, 
``The eICU Collaborative Research Database, a freely available multi-center database for critical care research,'' 
{\it Sci. Data}, vol. 5, p. 180178, 2018.

\bibitem{thoral2021amsterdamumcdb}
P. J. Thoral, J. M. Peppink, R. H. Driessen, E. J. G. Sijbrands, E. J. O. Kompanje, 
L. Kaplan, H. Bailey, {\it et al.}, 
``Sharing ICU patient data responsibly under the SCCM/ESICM Joint Data Science Collaboration: 
The Amsterdam University Medical Centers Database (AmsterdamUMCdb) example,'' 
{\it Crit. Care Med.}, vol. 49, no. 6, pp. e563--e577, 2021.

\bibitem{faltys2021hirid}
M. Faltys, M. Zimmermann, X. Lyu, M. Hüser, S. Hyland, G. Rätsch, and T. Merz, 
``HiRID, a high time-resolution ICU dataset,'' 
{\it PhysioNet}, version 1.1.1, 2021.


\bibitem{arnrich2024meds}
B. Arnrich, E. Choi, T. J. Pollard, M. Wornow, E. Steinberg, N. H. Shah, {\it et al.},
``Medical Event Data Standard (MEDS): Facilitating machine learning for health,''
in {\it Proc. Workshop on Trustworthy and Socially Responsible Machine Learning in Healthcare (TS4H) at ICLR}, 2024.

\bibitem{vandeWater2023yaib}
R. van de Water, M. Wornow, L. Atallah, B. Arnrich, and P. Rockenschaub,
``Yet Another ICU Benchmark: A flexible multi-center framework for clinical machine learning,''
arXiv:2306.05109, 2023.

\bibitem{guo2024ehrfm}
L. L. Guo, J. Fries, E. Steinberg, S. L. Fleming, K. E. Morse, C. Aftandilian, J. Posada,
N. H. Shah, and L. Sung,
``A multi-center study on the adaptability of a shared foundation model for electronic health records,''
{\it npj Digital Medicine}, vol. 7, no. 1, art. 171, 2024.

\bibitem{zhou2022mikgi}
D. Zhou, Z. Gan, X. Shi, A. Patwari, E. Rush, C.-L. Bonzel, {\it et al.},
``Multiview incomplete knowledge graph integration with application to cross-institutional EHR data harmonization,''
{\it J. Biomed. Inform.}, vol. 133, p. 104147, 2022.

\bibitem{shang2019gamenet}
J. Shang, C. Ma, J. Xiao, T. Sun, X. Li, and J. Jiang,
``GAMENet: Graph augmented memory networks for recommending medication combination,''
in {\it Proc. AAAI Conf. Artif. Intell.}, vol. 33, pp. 1126--1133, 2019.

\bibitem{hur2022descemb}
K. Hur, J. Lee, J. Oh, W. N. Price II, Y.-H. Kim, and E. Choi,
``Unifying heterogeneous electronic health records systems via text-based code embedding,''
in {\it Proc. ACM Conf. Health, Inference, and Learning (CHIL)}, 
Proc. Mach. Learn. Res., vol. 174, pp. 183--203, 2022.

\bibitem{devlin2019bert}
J. Devlin, M.-W. Chang, K. Lee, and K. Toutanova,
``BERT: Pre-training of deep bidirectional transformers for language understanding,''
in {\it Proc. NAACL-HLT}, pp. 4171--4186, 2019.

\bibitem{lee2020biobert}
J. Lee, W. Yoon, S. Kim, D. Kim, S. Kim, C. H. So, and J. Kang,
``BioBERT: A pre-trained biomedical language representation model for biomedical text mining,''
{\it Bioinformatics}, vol. 36, no. 4, pp. 1234--1240, 2020.

\bibitem{alsentzer2019clinicalbert}
E. Alsentzer, J. Murphy, W. Boag, W.-H. Weng, D. Jin, T. Naumann, and M. McDermott,
``Publicly available clinical BERT embeddings,''
in {\it Proc. 2nd Clinical Natural Language Processing Workshop}, pp. 72--78, 2019.

\bibitem{pires2019mbert}
T. Pires, E. Schlinger, and D. Garrette,
``How multilingual is multilingual BERT?,''
in {\it Proc. ACL}, pp. 4996--5001, 2019.

\bibitem{conneau2020xlmr}
A. Conneau, K. Khandelwal, N. Goyal, V. Chaudhary, G. Wenzek, F. Guzmán, E. Grave,
M. Ott, L. Zettlemoyer, and V. Stoyanov,
``Unsupervised cross-lingual representation learning at scale,''
in {\it Proc. ACL}, pp. 8440--8451, 2020.

\bibitem{geng2023kbioxlm}
L. Geng, X. Yan, Z. Cao, J. Li, W. Li, S. Li, X. Zhou, Y. Yang, and J. Zhang,
``KBioXLM: A knowledge-anchored biomedical multilingual pretrained language model,''
arXiv:2311.11564, 2023.

\bibitem{lange2022clinx}
L. Lange, H. Adel, J. Strötgen, and D. Klakow,
``CLIN-X: Pre-trained language models and a study on cross-task transfer for concept extraction in the clinical domain,''
{\it Bioinformatics}, vol. 38, no. 12, pp. 3267--3274, 2022.


\bibitem{johnson2021mimic}
A. E. W. Johnson, L. Bulgarelli, L. Shen, A. Gayles, A. Shammout, S. Horng, 
T. J. Pollard, S. Hao, B. Moody, B. Gow, L.-W. H. Lehman, L. A. Celi, and R. G. Mark,
``MIMIC-IV, a freely accessible electronic health record dataset,''
{\it Scientific Data}, vol. 10, no. 1, p. 1, 2023.

\bibitem{nwicu2024}
D. Moukheiber, W. Temps, B. Molgi, Y. Li, A. Lu, P. Nannapaneni, A. Chahin, 
S. Hao, F. Torres~Fabregas, L. A. Celi, A. Wong, M. Lloyd, X. Borrat~Frigola, 
H.-C. Lee, D. Schneider, T. J. Pollard, Y. Luo, A. Kho, and R. G. Mark,
``Northwestern ICU (NWICU) database (version 0.1.0),''
{\it PhysioNet}, 2024.

\bibitem{ehrshot2023}
M. Wornow, R. Thapa, E. Steinberg, J. Fries, and N. H. Shah,
``EHRSHOT: An EHR benchmark for few-shot evaluation of foundation models,''
in {\it Proc. NeurIPS Datasets and Benchmarks Track}, 2023.

\bibitem{yeche2021hirid}
H. Y\`{e}che, R. Kuznetsova, M. Zimmermann, M. H\"{u}ser, X. Lyu, M. Faltys, and G. R\"{a}tsch,
``HiRID-ICU-Benchmark --- A comprehensive machine learning benchmark on high-resolution ICU data,''
in {\it Proc. NeurIPS Datasets and Benchmarks Track}, 2021.

\bibitem{rodemund2023sicdb}
N. Rodemund, B. Wernly, C. Jung, C. Cozowicz, and A. Kok\"{o}fer,
``The Salzburg Intensive Care database (SICdb): an openly available critical care dataset,''
{\it Intensive Care Medicine}, vol. 49, no. 6, pp. 700--702, 2023.


\bibitem{kdigo2012aki}
Kidney Disease: Improving Global Outcomes (KDIGO) Acute Kidney Injury Work Group,
``KDIGO Clinical Practice Guideline for Acute Kidney Injury,''
{\it Kidney International Supplements}, vol.~2, no.~1, pp.~1--138, 2012.

\bibitem{kim2024remed}
J. Kim, C. Shim, B. S. K. Yang, C. Im, S. Y. Lim, H.-G. Jeong, and E. Choi,
``General-Purpose Retrieval-Enhanced Medical Prediction Model Using Near-Infinite History,''
in {\it Proc. 9th Machine Learning for Healthcare Conference (MLHC)}, 
Proc. Mach. Learn. Res., vol.~252, 2024.

\bibitem{lui2012langid}
M. Lui and T. Baldwin,
``langid.py: An off-the-shelf language identification tool,''
in {\it Proc. 50th Annu. Meeting Assoc. Comput. Linguistics: System Demonstrations},
Jeju, Korea, 2012, pp.~25--30.

\end{thebibliography}
\end{document}